\documentclass[lettersize,journal]{IEEEtran}
\usepackage{amsmath,amsfonts,amssymb}
\usepackage{algorithmic}
\usepackage{algorithm}
\usepackage{array}
\usepackage[caption=false,font=normalsize,labelfont=sf,textfont=sf]{subfig}
\usepackage{textcomp}
\usepackage{stfloats}
\usepackage{url}
\usepackage{hyperref}
\usepackage{verbatim}
\usepackage{graphicx}
\usepackage{booktabs}
\usepackage{multirow}
\usepackage{colortbl} 
\usepackage{xcolor}
\usepackage{cite}
\usepackage{enumitem}
\begin{document}

\title{Few-Shot Causal Representation Learning for Out-of-Distribution Generalization on Heterogeneous Graphs}

\author{Pengfei~Ding, Yan Wang,~\IEEEmembership{Senior Member,~IEEE,} Guanfeng Liu,~\IEEEmembership{Member,~IEEE,} Nan Wang, and ~Xiaofang~Zhou,~\IEEEmembership{Fellow,~IEEE}
\IEEEcompsocitemizethanks{
\IEEEcompsocthanksitem P. Ding, Y. Wang, G. Liu and N. Wang are with the School of Computing, Macquarie University, Sydney NSW 2109, Australia. E-mail: pengfei.ding2@students.mq.edu.au, \{guanfeng.liu, yan.wang\}@mq.edu.au, nan.wang12@students.mq.edu.au.
\IEEEcompsocthanksitem X. Zhou is with the Hong Kong University of Science and Technology, Clear Water Bay, Kowloon, Hong Kong. E-mail: zxf@cse.ust.hk.}
}

\markboth{Journal of \LaTeX\ Class Files,~Vol.~14, No.~8, August~2021}%
{Shell \MakeLowercase{\textit{et al.}}: A Sample Article Using IEEEtran.cls for IEEE Journals}


\maketitle

\begin{abstract}
Heterogeneous graph few-shot learning (HGFL) aims to address the label sparsity issue in heterogeneous graphs (HGs), which consist of various types of nodes and edges. The core concept of HGFL is to extract generalized knowledge from rich-labeled classes in a source HG, transfer this knowledge to a target HG to facilitate learning new classes with few-labeled training data, and finally make predictions on unlabeled testing data. Existing methods typically assume that the source HG, training data, and testing data all share the same distribution. However, in practice, distribution shifts among these three types of data are inevitable due to (1) the limited availability of the source HG that matches the target HG distribution, and (2) the unpredictable data generation mechanism of the target HG. Such distribution shifts can degrade the performance of existing methods, leading to a novel problem of out-of-distribution (OOD) generalization in HGFL. To address this challenging problem, we propose a \underline{C}ausal \underline{O}OD \underline{H}eterogeneous graph \underline{F}ew-shot learning model, namely COHF. In COHF, we first characterize distribution shifts in HGs with a structural causal model, where an invariance principle can be explored for OOD generalization. Then, following this invariance principle, we propose a new variational autoencoder-based heterogeneous graph neural network to mitigate the impact of distribution shifts. Finally, by integrating this network with a novel meta-learning framework, COHF effectively transfers knowledge to the target HG to predict new classes with few-labeled data. Extensive experiments on seven real-world datasets have demonstrated the superior performance of COHF over the state-of-the-art methods.
\end{abstract}


\section{Introduction}
\begin{figure}[t]
\setlength{\abovecaptionskip}{0cm} 
\setlength{\belowcaptionskip}{0cm} 
\centering
\scalebox{0.268}{\includegraphics{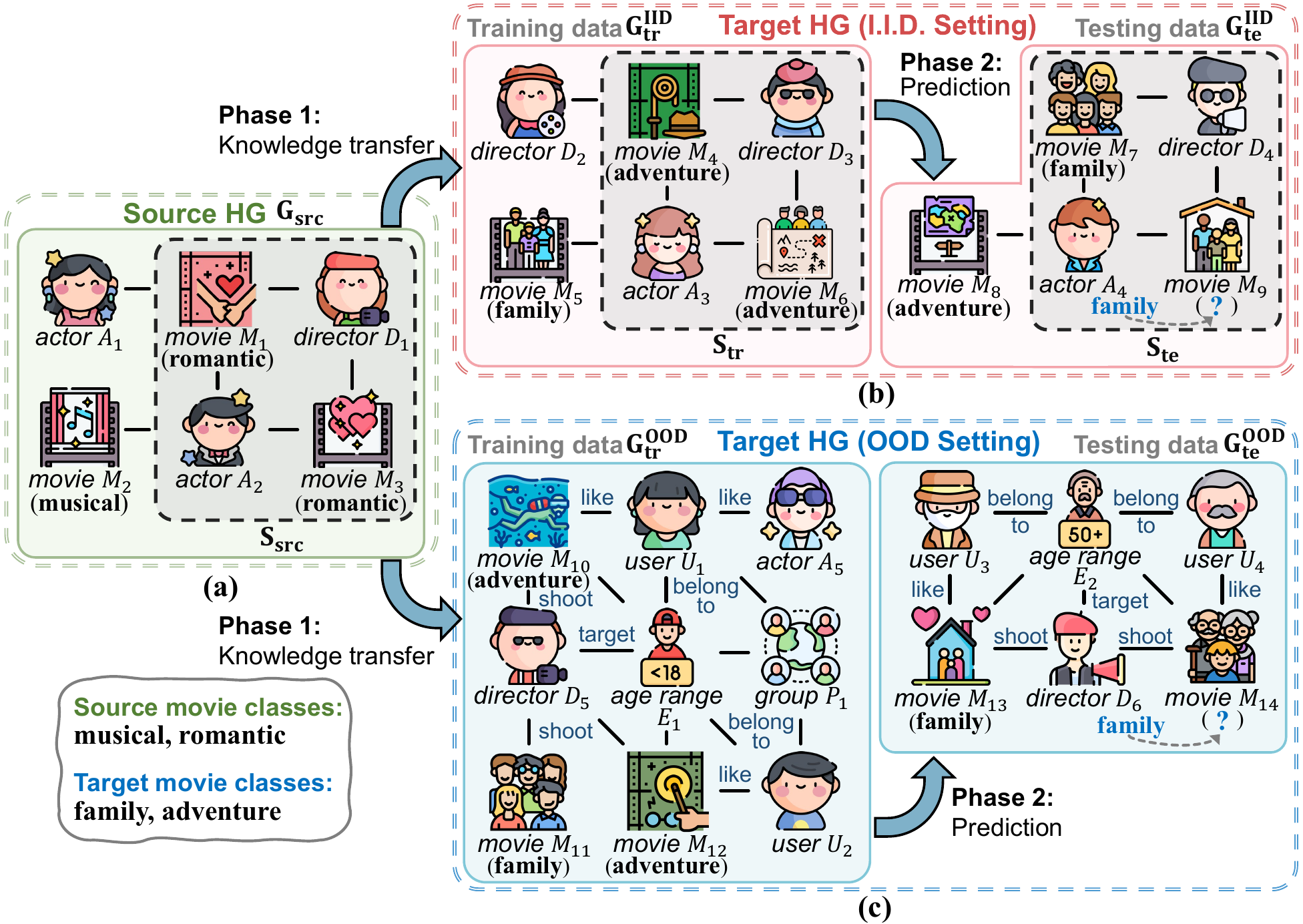}}
\caption{Heterogeneous graph few-shot scenarios in I.I.D. and OOD settings.}
\label{eg1}
\end{figure}

Heterogeneous graphs (HGs), consisting of various types of nodes and edges, have been widely used to model complex real-world systems \cite{shi2016survey}. Consider the HGs in Fig. \ref{eg1}, which model movie-related social networks. These HGs include multiple types of nodes (\textit{e.g.}, \texttt{user} and \texttt{actor}) and edges (\textit{e.g.}, \texttt{user} $\emph{U}_1$ “likes” \texttt{actor} $\emph{A}_5$, \texttt{director} $\emph{D}_4$ “shoots” \texttt{movie} $\emph{M}_{11}$). The variety of node and edge types necessitates substantial labeled data to learn effective representations on HGs for various downstream tasks (\emph{e.g.}, node classification) \cite{chen2023heterogeneous}. In real-world scenarios, label sparsity is a common issue because labeling nodes/edges/subgraphs with specific classes (\textit{e.g}., labeling protein subgraphs for therapeutic activity) demands considerable expertise and resources \cite{ju2023few}. To address this issue, {heterogeneous graph few-shot learning} (HGFL) has been developed, which aims to extract generalized knowledge (\textit{a.k.a.}, \textit{meta-knowledge}) from rich-labeled classes in a source HG, and then transfer this meta-knowledge to the target HG to facilitate learning new classes with few-labeled data \cite{ding2024few3}. Typical forms of meta-knowledge include node informativeness \cite{fang2024few}, heterogeneous information from shared node types \cite{zhuang2021hinfshot}, and semantic information of specific graph structures \cite{ding2023few}.


It is worth mentioning that existing HGFL studies generally assume that data is \emph{independently and identically distributed} (I.I.D.). This assumption posits that the rich-labeled data in the source HG, few-labeled training data in the target HG, and unlabeled testing data are all considered to come from the same distribution. The I.I.D. assumption simplifies both the problem modeling and the theoretical analysis, thereby easing the meta-knowledge extraction and the HGFL model design \cite{zhuang2021hinfshot,zhang2022hg}. For instance, in Fig. \ref{eg1}, under the I.I.D. setting, the source HG $\emph{G}_\emph{src}$, training data $\emph{G}^\emph{IID}_\emph{tr}$, and testing data $\emph{G}^\emph{IID}_\emph{te}$ exhibit similar subgraphs $\emph{S}_\emph{src}$, $\emph{S}_\emph{tr}$ and $\emph{S}_\emph{te}$, respectively. These subgraphs share a common circular structure that connects nodes of $<$\texttt{movie}, \texttt{director}, \texttt{movie}, \texttt{actor}$>$ in a cyclical pattern. Therefore, the general information regarding this circular structure can serve as meta-knowledge to facilitate prediction in $\emph{G}^\emph{IID}_\emph{te}$. Specifically, in $\emph{S}_\emph{src}$ of $\emph{G}_\emph{src}$, the fact that both \texttt{movie} $\emph{M}_1$ and $\emph{M}_3$ belong to the same class (\emph{romantic}) reveals knowledge that \texttt{movie} nodes within the circular structure tend to have similar classes. The generalizability of this knowledge can be validated in $\emph{S}_\emph{tr}$ of $\emph{G}^\emph{IID}_\emph{tr}$, where $\emph{M}_4$ and $\emph{M}_6$ are both \emph{adventure} movies. Consequently, this generalized knowledge can be used in $\emph{G}^\emph{IID}_\emph{te}$ to predict the class of \texttt{movie} nodes in $\emph{S}_\emph{te}$, \emph{i.e.}, $\emph{M}_9$ is likely to share the same class as $\emph{M}_7$ (\emph{action}).



However, in real-world scenarios, the I.I.D. assumption of existing studies is normally not satisfied. This is mainly due to (1) the limited availability of the source HG that closely matches the target HG distribution \cite{ding2023cross}, and (2) the unpredictable nature of data generation processes in the target HG \cite{bengio2019meta}. As a result, distribution mismatches (\textit{a.k.a.} \emph{distribution shifts}) are inevitable between the source HG and the target HG, leading to complex \emph{out-of-distribution} (OOD) environments in HGFL. As a new topic that remains unexplored in the existing literature, OOD environments in HGFL exhibit two unique characteristics as follows:

\noindent\textbf{1. Multi-level Characteristic:} Distribution shifts in HGFL can occur at three levels: \textit{feature-level} (\textit{e.g.}, variations in node/edge features), \textit{topology-level} (\textit{e.g.}, changes in graph size or other structural properties), and \textit{heterogeneity-level} (\textit{e.g.}, discrepancies in node/edge types). These different levels of distribution shifts often occur simultaneously. For instance, in Fig. \ref{eg1}(c), compared with the source HG $\emph{G}_\emph{src}$, the training data $\emph{G}^\emph{OOD}_\emph{tr}$ introduces new node types (\emph{e.g.}, \texttt{user} and \texttt{group}) and new edge types (\emph{e.g.}, “target” and “like”), leading to a heterogeneity-level shift between $\emph{G}_\emph{src}$ and $\emph{G}^\emph{OOD}_\emph{tr}$. Additionally, these new node and edge types contribute to a different graph structure and an increase in graph size, resulting in a topology-level shift between $\emph{G}_\emph{src}$ and $\emph{G}^\emph{OOD}_\emph{tr}$. 

\noindent\textbf{2. Phase-spanning Characteristic:} Distribution shifts in HGFL can occur in both the \emph{knowledge transfer} phase and the \emph{prediction} phase of the few-shot learning process. For example, in the OOD setting of Fig. \ref{eg1}, during the knowledge transfer phase, the introduction of new node and edge types in $\emph{G}^\emph{OOD}_\emph{tr}$ results in a heterogeneity-level shift between $\emph{G}_\emph{src}$ and $\emph{G}^\emph{OOD}_\emph{tr}$. In the prediction phase, differences in the \texttt{age range} (\emph{i.e.}, ‘below 18’ \emph{vs.} ‘over 50’) between $\emph{G}^\emph{OOD}_\emph{tr}$ and $\emph{G}^\emph{OOD}_\emph{te}$ introduce variations in \texttt{user} node features, yielding a feature-level shift between nodes in these two graphs. 

The multi-level and phase-spanning nature of OOD environments in HGFL not only hinders effective knowledge transfer from the source HG to the target HG, but also affects the prediction of testing data based on the features learned from training data. For instance, in Fig. \ref{eg1}(c), due to the topology-level shift between $\emph{G}_\emph{src}$ and $\emph{G}^\emph{OOD}_\emph{tr}$, the knowledge of the circular structure $\emph{S}_\emph{src}$ cannot be transferred to $\emph{G}^\emph{OOD}_\emph{tr}$. Moreover, the feature-level shift between $\emph{G}^\emph{OOD}_\emph{tr}$ and $\emph{G}^\emph{OOD}_\emph{te}$ can lead to incorrect predictions for \texttt{movie} nodes in $\emph{G}^\emph{OOD}_\emph{te}$. Therefore, it is crucial to investigate the consistency across source HG, training data, and testing data, which remains stable in contrast to distribution shifts. For instance, in Fig. \ref{eg1}(a), the knowledge that \texttt{movie} nodes connected to the same \texttt{director} node typically belong to the same class, can be inferred from $\emph{G}_\emph{src}$, \emph{e.g.}, \texttt{movie} $\emph{M}_1$ and $\emph{M}_3$ linked with \texttt{director} $\emph{D}_1$ are classified as \emph{romantic}. This knowledge can be further refined in the OOD setting of Fig. \ref{eg1}(c): \texttt{movie} nodes connected to the same \texttt{director} node may have the same class if they target the same \texttt{user} group, \emph{e.g.}, in $\emph{G}^\emph{OOD}_\emph{tr}$, $\emph{M}_{10}$ and $\emph{M}_{12}$ directed by $\emph{D}_5$ targeting \texttt{user} nodes ‘under 18’ fall into the same class (\emph{adventure}). This pattern remains consistent in $\emph{G}^\emph{OOD}_\emph{te}$, where $\emph{M}_{13}$ and $\emph{M}_{14}$ directed by $\emph{D}_6$ targeting \texttt{user} nodes aged ‘over 50’ belong to the same class (\emph{family}). This example illustrates the presence of consistency in OOD environments, and exploring such consistency is important for robust knowledge transfer and accurate predictions.

The above discussion reveals a novel problem of \emph{out-of-distribution generalization on heterogeneous graph few-shot learning}, which aims to handle distribution shifts not only between the source HG and the target HG, but also between the training and testing data in the target HG. To tackle this new problem, we have to face three key challenges as follows:

\noindent\textbf{CH1.} \emph{How to figure out the invariance principle in HGs to enable OOD generalization in heterogeneous graph few-shot learning?} 
The \emph{invariance principle} implies that model predictions, which focus only on the causes of the label, can remain consistent to a range of distribution shifts \cite{pearl2009causality,rojas2018invariant}. This principle forms the cornerstone of many studies that enable OOD generalization in regular Euclidean data (\textit{e.g.}, images) \cite{li2023rethinking}. While recent efforts have extended this principle to non-Euclidean data and addressed distribution shifts in graphs, they primarily focus on homogeneous graphs with a single type of nodes and a single type of edges \cite{liu2023good,fan2022debiasing}. However, exploring the invariance principle in heterogeneous graphs is non-trivial, because it requires extracting intricate semantics that remain unaffected by distribution shifts. Such semantics may comprise diverse node and edge types with complicated structural characteristics. Therefore, exploring a new invariance principle in HGs poses a significant challenge. 
 
\noindent\textbf{CH2.} \emph{How to achieve effective knowledge transfer from the source HG to the target HG in OOD environments?} Most HGFL methods transfer knowledge through shared node types or common graph structures between the source HG and the target HG \cite{fang2024few,zhang2022hg}. However, distribution shifts in HGFL can introduce heterogeneity and structural variances between these HGs, potentially leading to ineffective knowledge transfer or even negative transfer \cite{wang2022causal}. Considering the multi-level nature of these distribution shifts, developing a novel knowledge transfer framework is a substantial and complex challenge.



\noindent\textbf{CH3.} \emph{Given few-labeled data sampled from an OOD environment, how to effectively learn the class associated with these samples?} Existing few-shot learning methods are primarily designed for I.I.D. environments. These methods aim to derive comprehensive class information from limited samples \cite{kim2023task,huang2020graph}. However, OOD environments demand a focus on aspects of class information that remain invariant with distribution shifts, rather than attempting to capture all class information. This poses the challenge of identifying invariant class information with only a few labeled samples available.

To address the above three challenges, we propose a novel \underline{\textbf{C}}ausal \underline{\textbf{O}}OD \underline{\textbf{H}}eterogeneous graph \underline{\textbf{F}}ew-shot learning model, called COHF. To address \textbf{CH1}, we construct a structural causal model (SCM) to analyze the cause-effect relationships that influence the generation of node labels. Based on the SCM, we discover the invariance principle for OOD generalization in HGs: heterogeneous graph neural networks (HGNNs) remain invariant to distribution shifts by focusing only on environment-independent factors. To address \textbf{CH2}, we follow this principle and develop a new variational autoencoder-based HGNN module (VAE-HGNN). VAE-HGNN includes an encoder that infers invariant factors from the source HG, and a decoder that utilizes these invariant factors to predict labels in the target HG. To address \textbf{CH3}, we propose a novel meta-learning based module that transfers the knowledge of evaluating the richness of invariant factors from the source HG to the target HG. This module determines the importance of few-labeled samples to derive robust class representations for accurate prediction.

To the best of our knowledge, our work is the first to propose the novel problem of OOD generalization in heterogeneous graph few-shot learning and provide a solution for it. Our key contributions are as follows:
\begin{itemize}[leftmargin=*]
    \item We propose the COHF model, which performs novel causal modeling and causal inference to handle various distribution shifts in heterogeneous graph few-shot learning;
    \item We develop a novel heterogeneous graph neural network module that can extract invariant factors in HGs, and introduce a novel meta-learning module that can achieve effective knowledge transfer and robust predictive performance in OOD environments; 
    \item We conduct extensive experiments on seven real-world datasets. The experimental results illustrate that COHF outperforms the best-performing baselines by an average of 5.43\% in accuracy and 6.04\% in F1-score.
    

\end{itemize}

\section{Related Work}
\subsection{Heterogeneous Graph Representation Learning}
Heterogeneous graph representation learning aims to learn low-dimensional node embeddings for a variety of network mining tasks. Recently, heterogeneous graph neural networks (HGNNs) \cite{bing2023heterogeneous} have shown promising results in learning representations on HGs. HGNNs are primarily categorized into two groups: type-based and meta-path based. Type-based HGNNs model various types of nodes and edges directly. For example, HetSANN \cite{hong2020attention} incorporates type-specific graph attention layers to aggregate local information, while Simple-HGN \cite{lv2021we} adopts learnable edge type embeddings for edge attention. In contrast, meta-path based HGNNs focus on modeling meta-paths to extract hybrid semantics. For instance, HAN \cite{wang2019heterogeneous} employs a hierarchical attention mechanism to capture node-level importance and semantic-level importance. MAGNN \cite{fu2020magnn} proposes several meta-path encoders to encode comprehensive information along meta-paths. SeHGNN \cite{yun2019graph} adopts a transformer-based semantic fusion module to better utilize semantic information. However, these methods typically assume the same distribution between training and testing data, which limits their generalizability to OOD environments.

\subsection{Graph Few-shot Learning}
Graph few-shot learning combines few-shot learning with GNNs to address label sparsity issues. Most existing studies focus on homogeneous graphs \cite{zhang2022few}. For example, Meta-GNN \cite{zhou2019meta} combines GNNs with the MAML algorithm \cite{finn2017model} for effective graph learner initialization. G-Meta \cite{huang2020graph} utilizes meta-gradients from local subgraphs to update network parameters. Recent advancements extend few-shot learning to heterogeneous graphs. For instance, Meta-HIN \cite{fang2024few} employs multiple Bi-LSTMs to capture information from various node types and integrates the HGNN with a meta-learning framework for few-shot problems in a single HG. CrossHG-Meta \cite{zhang2022few2} and HG-Meta \cite{zhang2022hg} combine MAML with meta-path based HGNNs to enhance generalizability across HGs. However, these methods fail to address diverse distribution shifts in HGs, rendering them ineffective in OOD environments, \textit{e.g.}, if there is a heterogeneity-level shift between the source HG and the target HG, these methods cannot be generalized to the target HG due to differences in node types or meta-paths. CGFL \cite{ding2023cross} attempts to tackle the cross-heterogeneity challenge by simplifying heterogeneous information into two general patterns for knowledge transfer. However, this simplification may lead to the loss of specific node type information, limiting the capability to capture invariant heterogeneous information in OOD environments.

\subsection{OOD Generalization on Graphs} 
Some pioneering studies \cite{baranwal2021graph} have explored whether models trained on small graphs can effectively generalize to larger ones \cite{chuang2022tree}. Recently, researchers have extended OOD generalization methods, such as the invariance principle \cite{krueger2021out}, to explore graph structures that remain robust against distribution shifts. For instance, DIR \cite{wu2022discovering} divides a graph into causal and non-causal components using an edge threshold. CIGA \cite{chen2022learning} aims to maximize the agreement between the invariant portions of graphs with the same labels. CAL \cite{sui2022causal} reveals causal patterns and mitigates the confounding effects of shortcuts. DisC \cite{fan2022debiasing} investigates learning causal substructures within highly biased environments. However, most of these efforts focus on graph classification tasks in homogeneous graphs. Notably, there is a lack of research addressing more comprehensive distribution shifts in heterogeneous graphs.


\subsection{Domain-Invariant Feature Learning}
Domain-invariant feature learning aims to learn representations that are invariant across domains, which enables models trained on a labeled source domain (\textit{e.g.}, source graph) can be transferred to a target domain (\textit{e.g.}, target graph) without labeled samples (\textit{a.k.a.}, \textit{unsupervised domain adaption} (UDA)). While numerous studies focus on image and text data \cite{wang2019transferable,hao2021semi}, only a few methods have been developed for graph data. For instance, UDA-GCN \cite{wu2020unsupervised} achieves cross-domain node embedding by deceiving the domain discriminator. MuSDAC \cite{yang2021domain} focuses on heterogeneous graphs and utilizes multi-space alignment for domain adaptation across different semantic spaces. However, UDA methods assume that the target domain is entirely unlabeled. They primarily seek to minimize error in the source domain and bridge distribution gaps between domains through alignment techniques. In cases where the label distribution of the target domain is unknown, the performance of UDA methods cannot be guaranteed \cite{zhao2019learning}.



\section{Preliminaries}
\noindent\textbf{Heterogeneous Graph.} A heterogeneous graph is defined as $\emph{G}$ = $(\mathcal{V}, \mathcal{E}, \mathcal{S}_\emph{n}, \mathcal{S}_\emph{e})$, where $\mathcal{V}$ represents the set of nodes with a node type mapping function $\varphi(v): \mathcal{V} \mapsto \mathcal{S}_\emph{n}$, and $\mathcal{E}$ denotes the set of edges with an edge type mapping function $\psi(e): \mathcal{E} \mapsto \mathcal{S}_\emph{e}$. Here, $\mathcal{S}_\emph{n}$ and $\mathcal{S}_\emph{e}$ represent the sets of node types and edge types, respectively, with each node $v \in \mathcal{V}$ and each edge $e \in \mathcal{E}$ being associated with a specific type. Note that $|\mathcal{S}_\emph{n}| + |\mathcal{S}_\emph{e}|>2$ for heterogeneous graphs. Table \ref{tab_nota} summarizes notations frequently used in this paper for quick reference.

\noindent\textbf{Problem Formulation.} In this work, we focus on the problem of few-shot node classification in HGs under OOD environments. To mimic few-shot scenarios, we create two sets of few-shot tasks for the source HG and the target HG, named $\mathcal{T}_\emph{src}$ and $\mathcal{T}_\emph{tgt}$, respectively.

\begin{itemize}[leftmargin=*]
    \item \textbf{Input:} Source HG $\emph{G}_\emph{src}$ with its task set $\mathcal{T}_\emph{src}$. Target HG training data $\emph{G}_\emph{tgt}^\emph{tr}$ with its task set $\mathcal{T}_\emph{tgt}$. The distribution $\Psi(\emph{G}_\emph{src}) \neq \Psi(\emph{G}_\emph{tgt}^\emph{tr})$.
    \item \textbf{Output:} A model with good generalization ability to accurately predict the labels of nodes in the target HG testing data $\emph{G}_\emph{tgt}^\emph{te}$, where the distribution $\Psi(\emph{G}_\emph{tgt}^\emph{tr}) \neq \Psi(\emph{G}_\emph{tgt}^\emph{te})$. 
\end{itemize}


\begin{table}[]
\setlength{\abovecaptionskip}{0cm} 
\setlength{\belowcaptionskip}{0cm} 
\caption{Notations used in this paper.}
\label{tab_nota}
\centering
\begin{tabular}{cc}
\toprule[0.7pt]
\textbf{Notation}                      & \textbf{Explanation}                    \\ \midrule[0.5pt]
$\emph{G}$              & Heterogeneous graph      \\
$\mathcal{V}$         & Set of nodes          \\
$\mathcal{E}$         & Set of edges          \\
$\varphi(\cdot)$         & Node type mapping function  \\ 
$\psi(\cdot)$         & Edge type mapping function          \\
$\Phi$         & Single node type  \\
$\mathcal{S}_\emph{n}$         & Set of node types          \\
$\mathcal{S}_\emph{e}$         & Set of edge types          \\
$\tau$ = $(\tau_\emph{spt}, \tau_\emph{qry})$  & Single few-shot task\\
$\Psi(\emph{G})$          & Distribution of $\emph{G}$ \\
$p_{\theta}(\cdot)$   & Distribution parameterized by $\theta$ \\ 
$q_{\phi}(\cdot)$   & Variational distribution parameterized by $\phi$\\ 
$\textbf{rel}^\emph{c}$   & Set of common relations\\ 
$\textbf{rel}^\emph{u}$   & Set of unique relations\\

$\emph{a}_{uv}$  & Connectivity
between nodes $u$ and $v$ \\

$\emph{A}$  & Adjacency matrix \\
$\emph{G}_\emph{s}$         & Subgraph structure around the target node\\ 
$\mathbf{X}$     & Raw node features\\
$\mathbf{H}^{(l)}$  & Hidden embedding of the $l$-th layer\\
$\mathbf{h}_v$ & Hidden embedding of node $v$\\
$\mathbf{W}$  & Weight matrix\\
$\alpha, \gamma$  & Normalized attention weight\\
$\mathbf{proto}_c$         & Prototype embedding of class $c$ \\
\bottomrule[0.7pt]
\end{tabular}
\end{table}

\noindent\textbf{Few-shot Task Construction.} The set of $\emph{m}$ few-shot tasks on an HG, denoted as $\mathcal{T}$ = $\{\tau_1, \tau_2, \ldots, \tau_\emph{m}\}$, is constructed as follows: Under the $\emph{N}$-way $\emph{K}$-shot setting, each task $\tau$ = $(\tau_\emph{spt}, \tau_\emph{qry})$ is composed by first selecting $\emph{N}$ different classes $\emph{C}_{\tau}$ = $\{\emph{c}_1, \emph{c}_2, \ldots, \emph{c}_\emph{N}\}$ from the label space $\emph{C}$. Then, the support set $\tau_\emph{spt}$ = $\{\tau_{\emph{c}_1}, \tau_{\emph{c}_2}, \ldots, $ $\tau_{\emph{c}_\emph{N}}\}$ is formed by sampling $\emph{K}$ labeled nodes from each class, \textit{i.e.}, $\tau_{\emph{c}_\emph{i}}$ = $\{({v}_1, \emph{c}_\emph{i}), ({v}_2, \emph{c}_\emph{i}), \ldots, ({v}_\emph{K}, \emph{c}_\emph{i})\}$. The query set $\tau_\emph{qry}$ = $\{\tilde{\tau}_{\emph{c}_1}, \tilde{\tau}_{\emph{c}_2}, \ldots, \tilde{\tau}_{\emph{c}_\emph{N}}\}$ is created using the remaining nodes from each class, where $\tilde{\tau}_{\emph{c}_\emph{i}}$ = $\{(\tilde{{v}}_1, \emph{c}_\emph{i}), (\tilde{{v}}_2, \emph{c}_\emph{i}), \ldots, (\tilde{{v}}_\emph{K}, \emph{c}_\emph{i})\}$. 

For each task on the target HG ($\tau\in\mathcal{T}_\emph{tgt}$), the support set $\tau_\emph{spt}$ and query set $\tau_\emph{qry}$ are sampled from the training data $\emph{G}_\emph{tgt}^\emph{tr}$ and the testing data $\emph{G}_\emph{tgt}^\emph{te}$, respectively. After sufficient training iterations over all few-shot tasks in the source HG ($\mathcal{T}_\emph{src}$), the obtained model is expected to solve each few-shot task in the target HG ($\mathcal{T}_\emph{tgt}$) by performing $\emph{N}$-way classification in $\emph{G}_\emph{tgt}^\emph{te}$, utilizing only $\emph{K}$ labeled examples per class from $\emph{G}_\emph{tgt}^\emph{tr}$.

\section{Invariance Principle for OOD Generalization on Heterogeneous Graphs}
In this section, we first explore the label generation process in HGs from a causal perspective and introduce a novel structural causal model (SCM) to conceptualize this process. Next, we leverage this SCM to explore the invariance principle in HGs. This principle serves as the foundation for enabling OOD generalization in HGFL.

\subsection{Causal Perspective on Node Label Generation}
Considering a target node $v_t$ in an HG $\emph{G}$, we construct an SCM to model the label generation process of ${v}_t$ from a causal perspective (illustrated in Fig. \ref{scm}). The rationale behind this SCM is explained as follows:

\begin{itemize}[leftmargin=*]
    \item $\emph{E}_1$ represents the observed features of nodes and edges in $\emph{G}$. We consider that these features are shaped by the underlying environment behind the HG. Different environments can result in variations in $\emph{E}_1$. For instance, in social networks, when user communities originating from various environments (\textit{e.g.}, users from different countries or having different professions) are modeled as HGs, they generally exhibit distinct user characteristics and interaction attributes.
    
    \item $\emph{E}_2$ represents the unobserved features of nodes and edges in $\emph{G}$. We consider that these features are typically elusive and unaffected by environmental changes, \textit{e.g.}, the consistency of user behaviors across different social networks and the general tendencies of consumers to follow purchasing trends in e-commerce networks. Since $\emph{E}_1$ and $\emph{E}_2$ focus on different aspects of node and edge features and do not exhibit direct causal relationships between them, $\emph{E}_1$ and $\emph{E}_2$ are independent of each other. Moreover, both $\emph{E}_1$ and $\emph{E}_2$ include node-level and edge-level features but lack compound semantic information, making them global yet shallow in nature.

\begin{figure}[t]
\setlength{\abovecaptionskip}{0cm} 
\setlength{\belowcaptionskip}{0cm} 
\centering
\scalebox{0.39}{\includegraphics{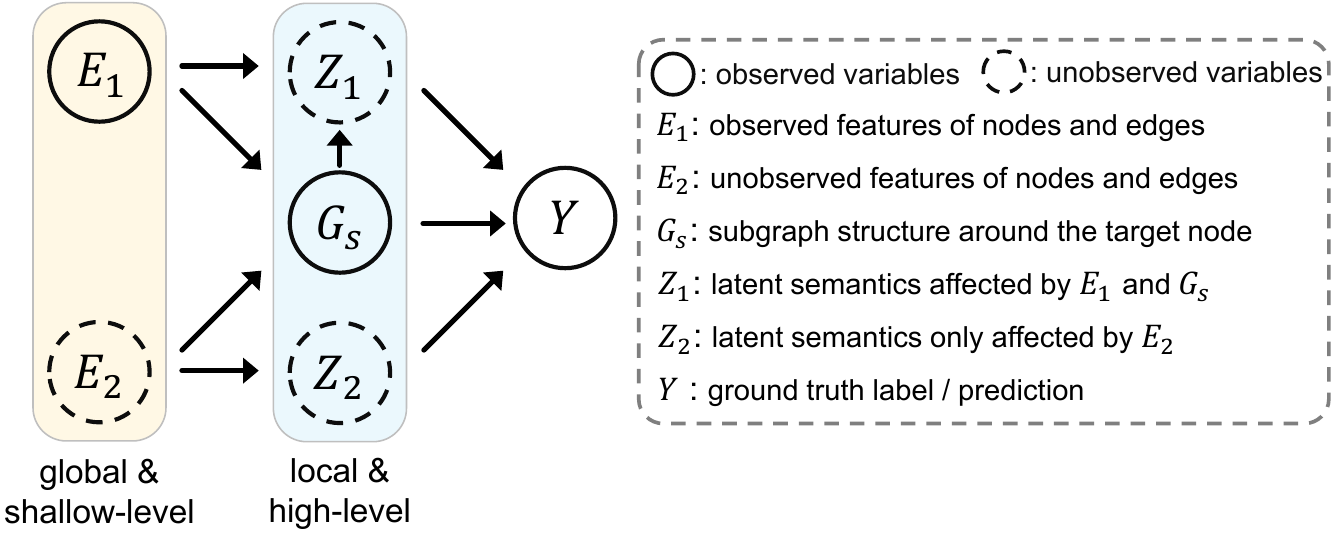}}
\caption{Structure causal model (SCM) for node label generation in HGs.}
\label{scm}
\end{figure}

    \item $\emph{G}_\emph{s}$ denotes the subgraph structure composed of the neighbors surrounding ${v}_t$. In particular, $\emph{G}_\emph{s}$ can be extracted by considering the $\emph{k}$-hop neighbors of ${v}_t$ in $\emph{G}$. 

    \item $\emph{Z}_1$ and $\emph{Z}_2$ represent semantics associated with ${v}_t$. $\emph{Z}_1$ includes semantics influenced by $\emph{E}_1$ and $\emph{G}_\emph{s}$, while $\emph{Z}_2$ represents semantics only affected by $\emph{E}_2$.


    \item $\emph{Y}$ denotes the category or label assigned to ${v}_t$.
    
    \item $(\emph{E}_1, \emph{E}_2)\rightarrow \emph{G}_\emph{s}$ means that the subgraph structure is determined by features of nodes and edges around ${v}_t$.


    \item $(\emph{E}_1, \emph{G}_\emph{s})$ $\rightarrow$ $\emph{Z}_1$ and $\emph{E}_2$ $\rightarrow$ $\emph{Z}_2$ denote that semantics are based on diverse features of nodes and edges. $\emph{Z}_1$ and $\emph{Z}_2$ are unobservable, and identifying them typically require domain-specific knowledge, \textit{e.g.}, researchers often utilize predefined meta-paths \cite{sun2011pathsim} and meta-graphs \cite{fang2019metagraph}, which consist of diverse node types, to extract semantic information. 
    
    
    \item $(\emph{Z}_1, \emph{Z}_2, \emph{G}_\emph{s})\rightarrow \emph{Y}$ means that the label of ${v}_t$ is determined by high-level semantics and the subgraph structure. 
\end{itemize}


\noindent\textbf{Connection with Existing Works and Novel Insights.} 
While our proposed SCM is the first causal model for node label generation in HGs, our analysis reveals that the SCM not only aligns with established paradigms in existing literature but also introduces some novel insights:

\begin{itemize}[leftmargin=*]
\item \textbf{Capturing semantics in HGs from shallow to high levels:} Directly capturing semantics within HGs is challenging \cite{wang2022survey}. Existing methods typically adopt a hierarchical learning strategy, leveraging diverse node types to capture complex heterogeneous information and reveal underlying semantics \cite{wang2019heterogeneous,fu2020magnn}. Differently, our SCM focuses on semantics in OOD environments and adopts a causality-driven perspective to derive semantics from shallow-level features.

\item \textbf{Leveraging semantics for node labeling:} Investigating node-related semantics is crucial for accurate node classification \cite{li2021leveraging,yang2021interpretable}. Existing methods focus on explicit semantics, which are shaped by identified meta-paths. In contrast, our SCM highlights the importance of latent semantics across various HGs, offering a new perspective on node labeling in OOD environments.

\item \textbf{Inferring labels from subgraphs:} Existing studies have proved that local subgraphs can retain substantial node-related properties for label prediction \cite{zhang2018link,huang2020graph}. Our SCM extends this concept by exploring the causal relationship between subgraphs and node labels, thereby achieving a more comprehensive understanding of node label generation.
\end{itemize}


\subsection{Exploring Invariance for OOD Generalization on HGs}
In on our proposed SCM, a distribution shift occurs when there are changes in at least one of the observed variables (\textit{i.e.}, $\emph{E}_1$, $\emph{G}_\emph{s}$ and $\emph{Y}$). In this work, we focus on the most common distribution shift scenario of graph OOD problems, \textit{i.e.}, \textit{covariate shift} \cite{fan2022debiasing,gui2022good}, where the input variables ($\emph{E}_1$ and $\emph{G}_\emph{s}$) change while the conditional distribution of the label (\textit{i.e.}, $\emph{P}(\emph{Y}| \emph{G}_\emph{s}, \emph{E}_1))$ stays the same. From a causal view, the distribution shift from $(\emph{E}_1$ = $\mathbf{e_1}, \emph{G}_\emph{s}$ = $\mathbf{g})$ to $(\emph{E}_1$ = $\mathbf{e^\prime_1}, \emph{G}_\emph{s}$ = $\mathbf{g^\prime})$ is formulated as an \emph{intervention} \cite{pearl2018book}, denoted as $\emph{do}(\emph{E}_1$ = $\mathbf{e^\prime_1}, \emph{G}_\emph{s}$ = $\mathbf{g^\prime})$. Therefore, OOD generalization in HGs can be modeled as a \emph{post-intervention inference} of node label probabilities $\emph{P}(\emph{Y}|\emph{do}(\emph{E}_1$ = $\mathbf{e^\prime_1}, \emph{G}_\emph{s}$ = $\mathbf{g^\prime}), \emph{E}_2)$.



Considering that the intervention $\emph{do}(\emph{E}_1$ = $\mathbf{e^\prime_1}, \emph{G}_\emph{s}$ = $\mathbf{g^\prime})$ only affects $\emph{Z}_1$, we can reuse the unaffected variables $\emph{E}_2$ and $\emph{Z}_2$ to perform post-intervention inference. Thus, the invariance principle for OOD generalization in HGs is that \emph{models are invariant to distribution shifts if they focus only on unaffected variables $\textit{E}_2$ and $\textit{Z}_2$}. Since $\emph{Z}_2$ is only determined by $\emph{E}_2$ and can be inferred from it, our main target is identifying $\emph{E}_2$. 


\begin{figure*}[t]
\setlength{\abovecaptionskip}{0cm} 
\setlength{\belowcaptionskip}{0cm} 
\centering
\scalebox{0.52}{\includegraphics{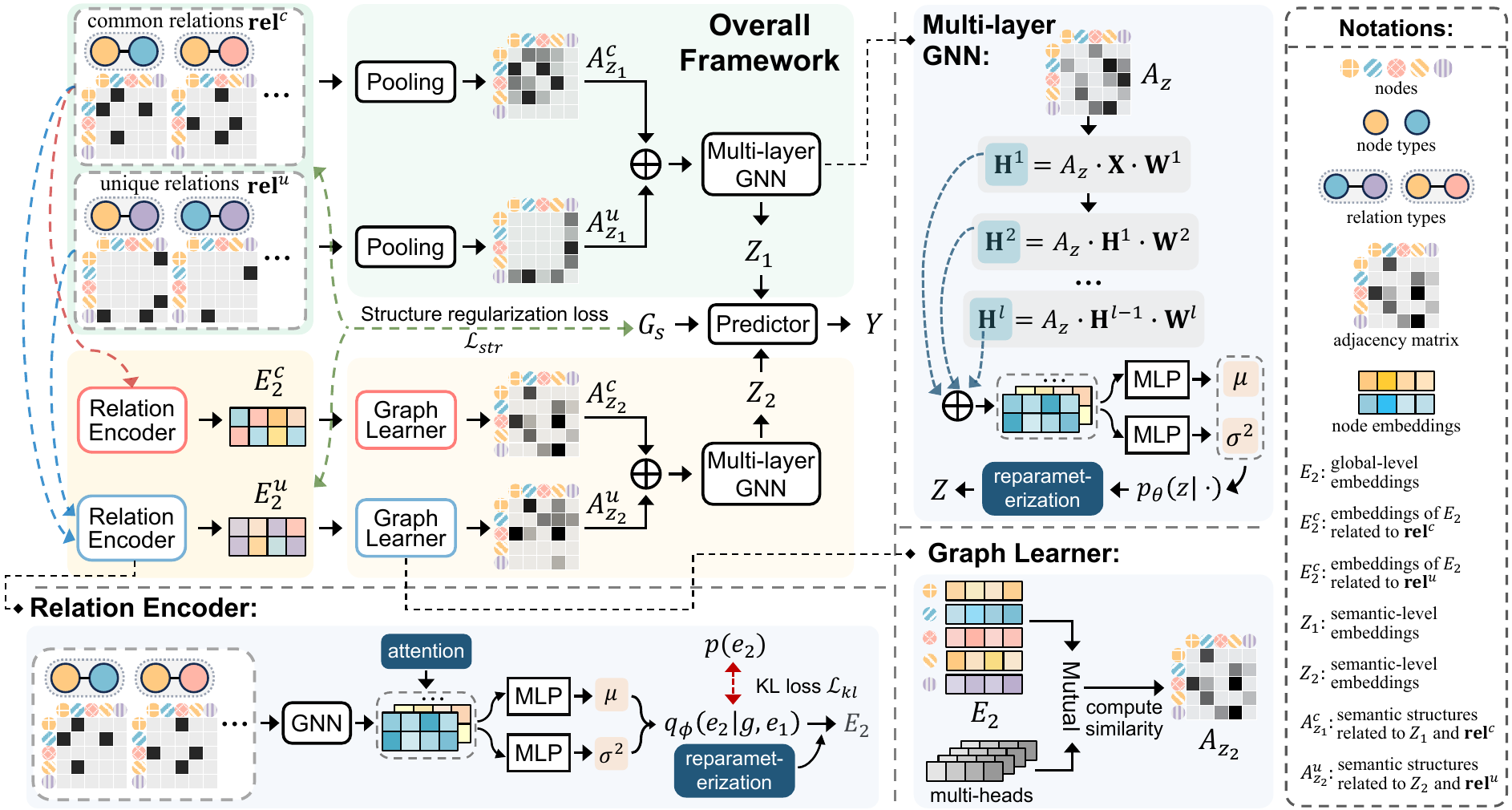}}
\caption{The overall architecture of VAE-HGNN.}
\label{model}
\end{figure*}

\section{Methodology}
\subsection{Overview}
In this section, we introduce our COHF model. COHF is built upon our SCM and follows the invariance principle to solve OOD problems in HGFL. Specifically, (1) To extract unaffected variables in the SCM, COHF proposes a variational autoencoder-based HGNN (VAE-HGNN) that models causal relationships within the SCM and applies causal inference on these relations to identify $\textit{E}_2$ and $\textit{Z}_2$. (2) To facilitate stable class learning from few-labeled samples, COHF proposes a node valuator module that evaluates the importance of each sample. By prioritizing samples that contain rich invariant features, COHF effectively captures label information that remains consistent despite distribution shifts. Overall, COHF extracts unaffected variables and focuses on invariant feature-rich samples to mitigate distribution shifts in HGFL, thereby facilitating effective few-shot learning in OOD environments.





\subsection{Variational Autoencoder-based HGNN}
Fig. \ref{model} illustrates the architecture of VAE-HGNN. VAE-HGNN includes three main modules to model the three unobservable variables (\textit{i.e.}, $\emph{E}_2$, $\emph{Z}_1$ and $\emph{Z}_2$) in the SCM: (1) The \emph{relation encoder} module infers $\emph{E}_2$ by leveraging the observed node features $\emph{E}_1$ and the subgraph structure $\emph{G}_\emph{s}$. (2) The \emph{multi-layer GNN} module infers $\emph{Z}_1$ based on $\emph{E}_1$ and $\emph{G}_\emph{s}$. (3) The \emph{graph learner} module infers $\emph{Z}_2$ from the previously inferred $\emph{E}_2$. Overall, VAE-HGNN can be regarded as an autoencoder, which consists of an encoder that infers $\emph{E}_2$, and a decoder that utilizes $\emph{E}_2$ with observed variables $\emph{E}_1$ and $\emph{G}_\emph{s}$ to infer $\emph{Y}$.

\noindent\textbf{Causal Inference for OOD Generalization in HGs.}
Before a distribution shift occurs, we represent all observed variables as ($\emph{E}_1$ = $\mathbf{e_1}$, $\emph{G}_\emph{s}$ = $\mathbf{g}$, $\emph{Y}$ = $\mathbf{y}$). Since the variable $\emph{E}_2$ = $\mathbf{e_2}$ remains unaffected to distribution shifts, our goal is to infer $\mathbf{e_2}$ conditional on these observed variables, \textit{i.e.}, to estimate the posterior distribution $p_{\theta}(\mathbf{e_2}|\mathbf{y}, \mathbf{e_1}, \mathbf{g})$, which is parameterized by $\theta$. However, this distribution is intractable due to the complexity of the joint distribution $p(\mathbf{y}, \mathbf{e_1}, \mathbf{g})$. To address this challenge, we resort to variational inference \cite{kingma2014semi,kingma2013auto} and introduce a variational distribution $q_{\phi}(\mathbf{e_2}|\mathbf{y}, \mathbf{e_1}, \mathbf{g})$, parameterized by $\phi$, to approximate the true posterior $p_{\theta}(\mathbf{e_2}|\mathbf{y}, \mathbf{e_1}, \mathbf{g})$.

\noindent\textbf{Learning Objective.} Following the VAE mechanism in \cite{simonovsky2018graphvae,kingma2013auto}, we optimize the model parameters $\theta$ and $\phi$ by maximizing the Evidence Lower BOund (ELBO), which is a lower bound for the log-likelihood of the observed data $\mathbf{y}$ and $\mathbf{g}$ conditioned on $\mathbf{e_1}$ (\textit{i.e.}, ${p}_\theta(\mathbf{y}, \mathbf{g}| \mathbf{e_1})$):
\begin{equation}
\label{elbo}
\begin{split}
 \log {p}_\theta(\mathbf{y}, \mathbf{g}| \mathbf{e_1}) &\geq \mathbb{E}_{{q}_\phi(\mathbf{e_2} | \mathbf{y}, \mathbf{g}, \mathbf{e_1})}\left[\log{p}_\theta(\mathbf{y}, \mathbf{g}|\mathbf{e_1},\mathbf{e_2})\right]-\\& \qquad \emph{KL}\left[{q}_\phi(\mathbf{e_2} | \mathbf{y}, \mathbf{g}, \mathbf{e_1}) \| {p}(\mathbf{e_2})\right]
\\&\triangleq-\mathcal{L}_\emph{ELBO},
\end{split}
\end{equation}
the detailed proof can be found in Appendix I.

The rationale behind Eq. (\ref{elbo}) is that minimizing the ELBO loss $\mathcal{L}_\emph{ELBO}$ corresponds to maximizing the log-likelihood of ${p}_\theta(\mathbf{y}, \mathbf{g}| \mathbf{e_1})$. By further decomposing $\mathcal{L}_\emph{ELBO}$, the learning objective can be transformed into the following three parts:

\begin{equation}
\begin{aligned}
\mathcal{L}_\emph{ELBO}= & \underbrace{-\mathbb{E}_{{q}_\phi(\mathbf{e_2} | \mathbf{y}, \mathbf{g}, \mathbf{e_1})}[\log{p}_{\theta}(\mathbf{y}|\mathbf{g},\mathbf{e_1},\mathbf{e_2})]}_{\text {Classification Loss } \mathcal{L}_\emph{cls}} \\
& \underbrace{-\mathbb{E}_{{q}_\phi(\mathbf{e_2} | \mathbf{y}, \mathbf{g}, \mathbf{e_1})}[\log{p}_{\theta}(\mathbf{g}|\mathbf{e_1},\mathbf{e_2})]}_{\text {Structure Regularization Loss } \mathcal{L}_\emph{str}} \\
& +\underbrace{\emph{KL}[{q}_\phi(\mathbf{e_2} | \mathbf{y}, \mathbf{g}, \mathbf{e_1}) \| {p}(\mathbf{e_2})]}_{\text {KL Loss }\mathcal{L}_\emph{kl}} ,
\end{aligned}
\end{equation}
where the first two terms account for the loss of node label prediction and subgraph structure regeneration, respectively, while the last term regularizes the Kullback-Leibler (KL) divergence between ${q}_\phi(\mathbf{e_2} | \mathbf{y}, \mathbf{g}, \mathbf{e_1})$ and ${p}(\mathbf{e_2})$. We assume that $\mathbf{e_2}$ is a $\emph{D}$-dimensional latent vector sampled from a standard Gaussian prior \cite{wang2020graph}, \textit{i.e.}, ${p}(\mathbf{e_2})$ = $\mathcal{N}(0, \mathbf{\mathbf{I}}_\emph{D})$. To compute $\mathcal{L}_\emph{ELBO}$, we design an encoder $f_\phi(\cdot)$ to model ${q}_\phi(\mathbf{e_2} | \mathbf{y}, \mathbf{g}, \mathbf{e_1})$; and two decoders $g_{\theta_1}(\cdot)$ and $g_{\theta_2}(\cdot)$ to model ${p}_{\theta}(\mathbf{y}|\mathbf{g},\mathbf{e_1},\mathbf{e_2})$ and ${p}_{\theta}(\mathbf{g}|\mathbf{e_1},\mathbf{e_2})$, respectively.




\noindent\textbf{Encoder Network.} 
To instantiate the posterior distribution ${q}_\phi$, we first make a further approximation from ${q}_\phi(\mathbf{e_2} | \mathbf{y}, \mathbf{g}, \mathbf{e_1})$ to ${q}_\phi(\mathbf{e_2} | \mathbf{g}, \mathbf{e_1})$. This approximation is known as the mean-field method, which is widely used in variational inference \cite{blei2017variational,lao2022variational}. Note that although $\mathbf{y}$ is not directly used as input for the encoder, it serves as the supervision information for the training process. Inspired by variational autoencoders \cite{kingma2013auto}, we define ${q}_\phi(\mathbf{e_2} | \mathbf{g}, \mathbf{e_1})$ as a multivariate Gaussian with a diagonal covariance structure, which is flexible and allows $\mathcal{L}_\emph{ELBO}$ to be computed analytically \cite{wang2020graph}. Accordingly, the encoder is designed as follows:
\begin{gather}
\mathbf{h}_{\mathbf{e_2}} = \operatorname{GNN}_{\phi}(\mathbf{g}, \mathbf{e_1}),
 \\
{q}_\phi(\mathbf{e_2} |\mathbf{g}, \mathbf{e_1})=\mathcal{N}\left(\mathbf{\mu}_{\phi}(\mathbf{h}_{\mathbf{e_2}}), \operatorname{diag}\{\sigma^2_{\phi}(\mathbf{h}_{\mathbf{e_2}})\}\right),\label{qe2}
\end{gather}
where $\operatorname{GNN}_{\phi}(\cdot)$ is a graph neural network that transforms the subgraph structure and node features into latent representations $\mathbf{h}_{\mathbf{e_2}}$. $\mu_{\phi}(\cdot)$ and $\sigma_{\phi}(\cdot)$ are multi-layer perceptrons (MLPs) that convert $\mathbf{h}_{\mathbf{e_2}}$ into the mean and standard deviations, respectively, to parameterize the distribution of ${q}_\phi(\mathbf{e_2} | \mathbf{g}, \mathbf{e_1})$. Since ${q}_\phi(\mathbf{e_2} | \mathbf{g}, \mathbf{e_1})$ takes on a Gaussian form, the reparameterization trick from variational autoencoders \cite{kingma2013auto} can be employed here to calculate derivatives \emph{w.r.t.} the encoder parameters. 

$\operatorname{GNN}_{\phi}(\cdot)$ is designed to derive node representations that are related to $\emph{E}_2$. Therefore, $\operatorname{GNN}_{\phi}(\cdot)$ should be capable of handling various types of $\emph{E}_1$ and $\emph{G}_\emph{s}$ that arise from diverse distribution shifts. Given that there are three levels of distribution shifts in OOD environments within HGFL: feature-level, topology-level, and heterogeneity-level, our design of $\operatorname{GNN}_{\phi}(\cdot)$ is guided by the following considerations:





\begin{itemize}[leftmargin=*]
\item Considering feature-level shifts, \textit{i.e.}, variations in node and edge features, since we focus on addressing the node classification problem, $\operatorname{GNN}_{\phi}(\cdot)$ transforms $\mathbf{g}$ and $\mathbf{e_1}$ into node-level hidden representations. These representations should be robust to variations in node features.

\item Considering topology-level shifts, $\operatorname{GNN}_{\phi}(\cdot)$ should be able to capture heterogeneous information that is not affected by changes in the graph structure. However, existing HGNNs typically capture heterogeneous information through specific graph patterns (\textit{e.g.}, meta-paths and meta-graphs \cite{fu2020magnn,fang2019metagraph}), which may not persist after topology-level shifts. Thus, we focus on the simple components of these patterns, namely, node-node relations (\textit{e.g.}, \texttt{paper}-\texttt{author}). While topology-level shifts might alter meta-paths and meta-graphs, the core information embedded in these fundamental relations remains unaffected \cite{shi2020rhine} and can be combined to capture higher-level semantics.



\item Considering heterogeneity-level shifts between two HGs $\emph{G}_1$ and $\emph{G}_2$, differences in node types between the two HGs can lead to different relation types. Therefore,we categorize relation types from the two HGs into two sets: $\textbf{rel}^\emph{c}$ that includes \textit{common} relation types in both HGs, and $\textbf{rel}^\emph{u}$ that contains \textit{unique} relation types for each HG. Formally, $\textbf{rel}^\emph{c}$ = $\{(\Phi_{i}, \Phi_{j})|(\Phi_{i}, \Phi_{j}) \in \mathcal{S}_\emph{n}^1, (\Phi_{i}, \Phi_{j}) \in \mathcal{S}_\emph{n}^2\}$, where $\Phi_{i}$ and $\Phi_{j}$ are two node types in $\emph{G}_1$ and $\emph{G}_2$, respectively, and $\mathcal{S}_\emph{n}^1$ and $\mathcal{S}_\emph{n}^2$ are the node type sets of $\emph{G}_1$ and $\emph{G}_2$, respectively. $\textbf{rel}^\emph{u}_1$ = $\{(\Phi_{i}, \Phi_{j})|(\Phi_{i}, \Phi_{j}) \in \mathcal{S}_\emph{n}^1, (\Phi_{i}, \Phi_{j}) \notin \mathcal{S}_\emph{n}^2\}$ is the unique relation set in $\emph{G}_1$. Thus, a heterogeneity-level shift from $\emph{E}_1$ = $\mathbf{e_1}$ to $\emph{E}_1$ = $\mathbf{e^\prime_1}$ implies that $\textbf{rel}^\emph{c}(\mathbf{e_1})$ = $\textbf{rel}^\emph{c}(\mathbf{e^\prime_1})$ while $\textbf{rel}^\emph{u}(\mathbf{e_1}) \cup \textbf{rel}^\emph{u}(\mathbf{e^\prime_1}) \neq \emptyset$.


\end{itemize}




Based on the above analysis, we propose two subgraph-based relation encoders, $f_\phi^\emph{c}(\cdot)$ and $f_\phi^\emph{u}(\cdot)$, to learn the distributions of $\mathbf{e_2}$ related to $\textbf{rel}^\emph{c}$ and $\textbf{rel}^\emph{u}$, denoted as ${q}_\phi(\mathbf{e_2^c} |\mathbf{g}, \mathbf{e_1})$ and ${q}_\phi(\mathbf{e_2^u} |\mathbf{g}, \mathbf{e_1})$, respectively. Specifically, we first extract $\emph{n}_\emph{k}$-hop neighbors of the target node $v_t$ to construct the subgraph structure $\emph{G}_\emph{s}$. Then, we use the adjacency matrices of $\textbf{rel}^\emph{c}$ and $\textbf{rel}^\emph{u}$ in $\emph{G}_\emph{s}$, denoted as $\mathbf{A}^\emph{c}$ and $\mathbf{A}^\emph{u}$ respectively, as the input for $f_\phi^\emph{c}(\cdot)$ and $f_\phi^\emph{u}(\cdot)$. Formally, $\mathbf{A}^\emph{c}$ = $\{\emph{A}_{i,j}|(\Phi_{i}, \Phi_{j})\in \textbf{rel}^\emph{c}\}$ and $\mathbf{A}^\emph{u}$ = $\{\emph{A}_{i,j}|(\Phi_{i}, \Phi_{j})\in \textbf{rel}^\emph{u}\}$, where the adjacency matrix $\emph{A}_{i,j}$ = $\{\emph{a}_\emph{mn} | \emph{m}, \emph{n} \in \mathcal{V}_\emph{s}, \varphi(\emph{m})$ = $\Phi_{i}, \varphi(\emph{n})$ = $\Phi_{j}\}$, here $\emph{a}_\emph{mn}$ is the connectivity between nodes $\emph{m}$ and $\emph{n}$. 

For learning features of $\mathbf{e_2}$ from $\textbf{rel}^\emph{c}$ and $\textbf{rel}^\emph{u}$, we assume that both relations contain features that remain unaffected by distribution shifts. For $\textbf{rel}^\emph{c}$, the unaffected features are inherent within the relations themselves. In contrast, for $\textbf{rel}^\emph{u}$, the unaffected features arise from the interactions between $\textbf{rel}^\emph{c}$ and $\textbf{rel}^\emph{u}$. These interactions essentially reflect the mechanism by which common relations interact with unique relations. Specifically, for the common relation encoder $f_\phi^\emph{c}(\cdot)$, we first employ a GNN to learn the representation of a node $v$ under the $i$-th relation type $\emph{r}_i\in\textbf{rel}^\emph{c}$:
\begin{equation}
\label{hic}
\mathbf{h}^\emph{c}_{v\emph{-}i} = \operatorname{GNN}_\emph{c}\left(\emph{A}_i^\emph{c}, \mathbf{x}_v\right),
\end{equation}
where $\emph{A}_i^\emph{c}\in \mathbf{A}^\emph{c}$ is the adjacency matrix of $\emph{r}_i$, $\mathbf{x}_v$ is the original feature of node $v$. Then, we adopt an attention mechanism to aggregate information from different relation types in $\textbf{rel}^\emph{c}$:
\begin{gather}
\alpha^{\emph{c}}_{v\emph{-}i} = \frac{\exp\left(\operatorname{ReLU}\left(\mathbf{a}_\emph{c}^{\mathsf{T}}\cdot\ \mathbf{h}_{v\emph{-}i}^{\emph{c}} \right)\right)}{\sum\nolimits^{|\textbf{rel}^\emph{c}|}_{{j}=1}\exp\left(\operatorname{ReLU}\left(\mathbf{a}_\emph{c}^{\mathsf{T}}\cdot\mathbf{h}_{v\emph{-}j}^{\emph{c}} \right)\right)},
 \\
\mathbf{h}^{\emph{c}}_v=\sum\nolimits^{|\textbf{rel}^\emph{c}|}_{{i}=1} \alpha^{\emph{c}}_{v\emph{-}i} \cdot \mathbf{h}_{v\emph{-}i}^{\emph{c}}, 
\end{gather}
where $\alpha^{\emph{c}}_{v\emph{-}i}$ denotes the importance of the ${i}$-th type in $\textbf{rel}^\emph{c}$, $\mathbf{a}_\emph{c}$ represents the attention parameter shared among all common relation types. The unique relation encoder $f_\phi^\emph{u}(\cdot)$ has the same structure as $f_\phi^\emph{c}(\cdot)$ but processes different inputs. Specifically, when learning the representations of nodes under a unique relation, we consider integrating the interaction between the unique relation and common relations $\mathbf{A}^\emph{c}$ into another GNN:
\begin{equation}
    \mathbf{h}^\emph{u}_{v\emph{-}i} = \operatorname{GNN}_\emph{u}\left(\emph{A}^\emph{u}_i \cdot {f}^\emph{c}_\emph{pool}\left(\mathbf{A}^\emph{c}\right), \mathbf{x}_v\right),
\end{equation}
where $\emph{A}_i^\emph{u}$ is the adjacency matrix of the $i$-th unique relation, ${f}^\emph{c}_\emph{pool}(\cdot)$ is a pooling function (\textit{e.g.}, $\emph{max-pool}(\cdot)$, $\emph{sum-pool}(\cdot)$, and $\emph{mean-pool}(\cdot)$). Then, in the attention mechanism, we concatenate the representations of common relations and unique relations to compute the importance of each unique relation:
\begin{gather}
\alpha^{\emph{u}}_{v\emph{-}i} = \frac{\exp\left(\operatorname{ReLU}\left(\mathbf{a}_\emph{u}^{\mathsf{T}}\cdot\left[ \mathbf{h}_{v\emph{-}i}^{\emph{u}}\|\mathbf{h}_v^{\emph{c}}\right] \right)\right)}{\sum\nolimits^{|\textbf{rel}^\emph{u}|}_{{j}=1}\exp\left(\operatorname{ReLU}\left(\mathbf{a}_\emph{u}^{\mathsf{T}}\cdot\left[ \mathbf{h}_{v\emph{-}j}^{\emph{u}}\|\mathbf{h}_v^{\emph{c}}\right] \right)\right)},
 \\
\mathbf{h}^{\emph{u}}_v=\sum\nolimits^{|\textbf{rel}^\emph{u}|}_{{i}=1} \alpha^{\emph{u}}_{v\emph{-}i} \cdot \mathbf{h}_{v\emph{-}i}^{\emph{u}}, 
\end{gather}
where $\alpha^{\emph{u}}_{v\emph{-}i}$ denotes the importance of the ${i}$-th type in $\textbf{rel}^\emph{u}$, and $\mathbf{a}_\emph{u}$ represents the trainable attention parameter shared among all unique relation types. Finally, following Eq. (\ref{qe2}), we input $\mathbf{h}^{\emph{c}}$ and $\mathbf{h}^{\emph{u}}$ into different MLPs to compute the distributions of ${q}_\phi(\mathbf{e_2^c} |\mathbf{g}, \mathbf{e_1})$ and ${q}_\phi(\mathbf{e_2^u} |\mathbf{g}, \mathbf{e_1})$, respectively.



\noindent\textbf{Decoder Networks.} We propose two decoder networks $g_{\theta_1}(\cdot)$ and $g_{\theta_2}(\cdot)$ to model ${p}_{\theta}(\mathbf{y}|\mathbf{g},\mathbf{e_1},\mathbf{e_2})$ and ${p}_{\theta}(\mathbf{g}|\mathbf{e_1},\mathbf{e_2})$, respectively. For $g_{\theta_1}(\cdot)$, we first factorize ${p}_{\theta}(\mathbf{y}|\mathbf{g},\mathbf{e_1},\mathbf{e_2})$ based on our proposed SCM as follows:
\begin{equation}
\label{factorize}
\begin{split}
{p}_{\theta}(\mathbf{y}|\mathbf{g},\mathbf{e_1},\mathbf{e_2}) =&\int_{\mathbf{z_1}}\int_{\mathbf{z_2}} {p}_{\theta}(\mathbf{z_1} | \mathbf{g},\mathbf{e_1}) {p}_{\theta}(\mathbf{z_2} | \mathbf{e_2})\\&{p}_{\theta}(\mathbf{y}|\mathbf{g}, \mathbf{z_1}, \mathbf{z_2})d\mathbf{z_1}d\mathbf{z_2}.
\end{split}
\end{equation}




Based on the above factorization result, the core idea of $g_{\theta_1}(\cdot)$ is to utilize node features $\emph{E}_1$ = $\mathbf{e_1}$ and $\emph{E}_2$ = $\mathbf{e_2}$ to capture semantics $\emph{Z}_1$ = $\mathbf{z_1}$ and $\emph{Z}_2$ = $\mathbf{z_2}$, and then employ $\mathbf{z_1}$ and $\mathbf{z_2}$ together with $\mathbf{g}$ to infer the label $\mathbf{y}$. Specifically, we consider $\emph{Z}_1$ and $\emph{Z}_2$ as semantic-level features derived from specific graph structures determined by $\mathbf{e_1}$ and $\mathbf{e_2}$, respectively. Therefore, we first evaluate the connectivity of nodes influenced by $\mathbf{e_1}$ and $\mathbf{e_2}$, generating distinct graph structures that represent different semantics. Then, we implement representation learning within these specific graph structures. Next, we will detail the instantiation of each factorized distribution.

\noindent\textbf{Instantiating ${p}_{\theta}(\mathbf{z_1} | \mathbf{g},\mathbf{e_1})$ with Multi-layer GNN:} We utilize various meta-paths in the subgraph $\emph{G}_\emph{s}$ = $\mathbf{g}$ to capture semantics derived from $\emph{E}_1$ = $\mathbf{e_1}$. A meta-path consists of diverse node types and conveys specific semantics \cite{fu2020magnn}. For example, the meta-path \texttt{author}-\texttt{paper}-\texttt{author} indicates author collaboration. Existing studies typically require manually defined meta-paths \cite{sun2011pathsim}. However, in scenarios involving distribution shifts and few-shot learning, setting and sampling meaningful and significant meta-paths is challenging \cite{chen2017task}. Therefore, we propose a multi-layer GNN to automatically capture both short and long meta-paths across various relations. For the adjacency matrices of common relations ($\mathbf{A}^\emph{c}$) and unique relations ($\mathbf{A}^\emph{u}$), each relation matrix can be considered as a 1-length meta-path graph matrix. We first combine $\mathbf{A}^\emph{c}$ and $\mathbf{A}^\emph{u}$ to construct a weighted 1-length meta-path graph matrix:
\begin{equation}
\label{az1}
\emph{A}_{\emph{z}_1} = \beta_{\emph{z}_1}\times f^{\emph{z}_1}_\emph{pool}(\mathbf{A}^\emph{c}) + (1-\beta_{\emph{z}_1})\times f^{\emph{z}_1}_\emph{pool}(\mathbf{A}^\emph{u}),
\end{equation}
where $\beta_{\emph{z}_1}$ is a trainable smoothing parameter. To extract longer meta-paths, successive multiplications can be performed on the matrix $\emph{A}_{\emph{z}_1}$. For example, raising $\emph{A}_{\emph{z}_1}$ to the second power automatically incorporates information about 2-length meta-paths. This approach is commonly employed in existing meta-path extraction methods \cite{yun2019graph,gasteiger2018predict}. Therefore, to capture more diverse and longer meta-paths, we extend this approach to a network structure with $l$ layers as follows:
\begin{equation}
\begin{split}
\label{az1-l}
\mathbf{H}^{(l)}_{\emph{z}_1} & =\emph{A}_{\emph{z}_1} \cdot \mathbf{H}^{(l-1)}_{\emph{z}_1} \cdot \mathbf{W}^{(l)}_{\emph{z}_1} \\
& =\underbrace{\emph{A}_{\emph{z}_1} \cdots(\emph{A}_{\emph{z}_1}}_l \cdot \mathbf{X} \cdot \underbrace{\mathbf{W}^{(1)}_{\emph{z}_1}) \cdots \mathbf{W}^{(l)}_{\emph{z}_1}}_l \\
& =\emph{A}_{\emph{z}_1}^l \cdot \mathbf{X} \cdot \prod_{i=1}^{l}
\mathbf{W}^{i}_{\emph{z}_1},
\end{split}
\end{equation}
where $\mathbf{H}^{(l)}_{\emph{z}_1}$ is the hidden embedding of the $l$-th layer, $\mathbf{X}$ is the node feature matrix of $\emph{G}_\emph{s}$, and $\mathbf{W}^{(l)}_{\emph{z}_1}$ is the learnable weight matrix that evaluates the influence of meta-paths with $l$-length on the embedding. Finally, we fuse the outputs of all layers to capture the meta-path information of different lengths across various relations as follows:
\begin{equation}
\label{hz1}
\begin{split}
\mathbf{H}_{\emph{z}_1} & =\frac{1}{l} \sum_{i=1}^l \mathbf{H}^{(i)}_{\emph{z}_1} \\
& =\frac{1}{l} \sum_{i=1}^l \emph{A}_{\emph{z}_1}^l \cdot \mathbf{H}^{(i-1)}_{\emph{z}_1} \cdot \mathbf{W}^{(i)}_{\emph{z}_1},
\end{split}
\end{equation}
where $\mathbf{H}^{(0)}_{\emph{z}_1}$ is the node feature matrix $\mathbf{X}$. Then, we adopt two MLPs $\mu_{\theta_{\emph{z}_1}}(\cdot)$ and $\sigma_{\theta_{\emph{z}_1}}(\cdot)$ to estimate the parameters of ${p}_{\theta}(\mathbf{z_1} | \mathbf{g},\mathbf{e_1})$ as follows:
\begin{equation}
    {p}_{\theta}(\mathbf{z_1} | \mathbf{g},\mathbf{e_1})=\mathcal{N}\left(\mu_{\theta_{\emph{z}_1}}(\mathbf{H}_{\emph{z}_1}), \operatorname{diag}\{\sigma^2_{\theta_{\emph{z}_1}}(\mathbf{H}_{\emph{z}_1})\}\right).
    \label{qe3}
\end{equation}

\noindent\textbf{Instantiating ${p}_{\theta}(\mathbf{z_2} | \mathbf{e_2})$ with Graph Learner:} We first sample $\mathbf{e_2^c}$ and $\mathbf{e_2^u}$ from the distributions learned by the encoder networks, \textit{i.e.}, ${q}_\phi(\mathbf{e_2^c}|\mathbf{y}, \mathbf{g}, \mathbf{e_1})$ and ${q}_\phi(\mathbf{e_2^u}|\mathbf{y}, \mathbf{g}, \mathbf{e_1})$, respectively. Since $\mathbf{z_2}$ does not depend on the subgraph structure $\mathbf{g}$, we reconstruct the node connectivity regarding common and unique relations based on the node features of $\mathbf{e_2^c}$ and $\mathbf{e_2^u}$. Then, we leverage a GNN module to capture semantics based on the reconstructed graph structure. Specifically, we propose two graph learner modules to construct adjacency matrices for $\mathbf{e_2^c}$ and $\mathbf{e_2^u}$, denoted as $\emph{A}_{\emph{z}_2}^\emph{c}$ and $\emph{A}_{\emph{z}_2}^\emph{u}$, respectively. 
In each graph learner module, a multi-head weighted cosine similarity learning method is used to compute the connectivity of nodes:
\begin{equation}
\label{az2}
\emph{a}_{uv}^\emph{c}=\frac{1}{\emph{n}_\emph{att}} \sum_{i=1}^{\emph{n}_\emph{att}} \cos \left(\mathbf{w}^{\emph{c}}_{i} \cdot \mathbf{h}^{{{\emph{e}_2}}}_{u}, \mathbf{w}^\emph{c}_i \cdot \mathbf{h}^{{\emph{e}_2}}_{v}\right),
\end{equation}
where $\mathbf{h}^{{{\emph{e}_2}}}_{u}$ and $\mathbf{h}^{{{\emph{e}_2}}}_{v}$ are node representations in $\mathbf{e_2^c}$ for a node pair $(u,v)$, the set $\{\mathbf{w}^\emph{c}_i\}^{\emph{n}_\emph{att}}_{i=1}$ consists of ${\emph{n}_\emph{att}}$ independent learnable weight vectors that perform similarly as attention mechanisms \cite{vaswani2017attention}. After obtaining $\emph{A}_{\emph{z}_2}^\emph{c}$ and $\emph{A}_{\emph{z}_2}^\emph{u}$, we adopt a trainable parameter $\beta_{\emph{z}_2}$ to aggregate them and use Multi-layer GNN to learn the semantic-level representation of $\emph{Z}_2$:
\begin{gather}
\emph{A}_{\emph{z}_2} = \beta_{\emph{z}_2}\times \emph{A}_{\emph{z}_2}^\emph{c} + (1-\beta_{\emph{z}_2})\times \emph{A}_{\emph{z}_2}^\emph{u}, \label{z2}
\\
\mathbf{H}_{\emph{z}_2} = \operatorname{GNN}_{\emph{z}_2}\left(\emph{A}_{\emph{z}_2}, \mathbf{X}\right).\label{h2}
\end{gather}

Finally, two MLPs $\mu_{\theta_{\emph{z}_2}}(\cdot)$ and $\sigma_{\theta_{\emph{z}_2}}(\cdot)$ are used to estimate the parameters of ${p}_{\theta}(\mathbf{z_2} |\mathbf{e_2})$:
\begin{equation}
    {p}_{\theta}(\mathbf{z_2} |\mathbf{e_2})=\mathcal{N}\left(\mu_{\theta_{\emph{z}_2}}(\mathbf{H}_{\emph{z}_2}), \operatorname{diag}\{\sigma^2_{\theta_{\emph{z}_2}}(\mathbf{H}_{\emph{z}_2})\}\right).
    \label{qe4}
\end{equation}
\noindent\textbf{Approximating ${p}_{\theta}(\mathbf{y}|\mathbf{g}, \mathbf{z_1}, \mathbf{z_2})$:} Even if we obtain the estimations of ${p}_{\theta}(\mathbf{z_1} | \mathbf{g},\mathbf{e_1})$ and ${p}_{\theta}(\mathbf{z_2} | \mathbf{e_2})$, the calculation of ${p}_{\theta}(\mathbf{y}|\mathbf{g}, \mathbf{z_1}, \mathbf{z_2})$ (\textit{i.e.}, Eq. (\ref{factorize})) remains challenging due to the costly integral over the latent variables $\mathbf{z_1}$ and $\mathbf{z_2}$. To achieve higher efficiency, we resort to Monte Carlo (MC) sampling, which samples $\mathbf{z_1}$ and $\mathbf{z_2}$ to approximate ${p}_{\theta}(\mathbf{y}|\mathbf{g}, \mathbf{z_1}, \mathbf{z_2})$: 
\begin{equation}
\label{approx}
    {p}_{\theta}(\mathbf{y}|\mathbf{g}, \mathbf{z_1}, \mathbf{z_2}) \approx \frac{1}{\textit{L}} \frac{1}{\textit{M}} \sum_{i=1}^\textit{L} \sum_{j=1}^\textit{M} {p}_{\theta}\left(\mathbf{y} | \mathbf{g}, \mathbf{z}_1^i, \mathbf{z}_2^j\right),
\end{equation}
where $\textit{L}$ and $\textit{M}$ are the sample numbers; $\mathbf{z}_1^i$ and $\mathbf{z}_2^j$ are drawn from ${p}_{\theta}(\mathbf{z_1} | \mathbf{g},\mathbf{e_1})$ and ${p}_{\theta}(\mathbf{z_2} | \mathbf{e_2})$, respectively. Nevertheless, Eq. (\ref{approx}) is still computationally costly due to calculating the conditional probability many times (\textit{i.e.}, $\textit{L}\times\textit{M}$). We thus further conduct a widely used approximation \cite{gao2017bounds,wang2020visual}, which is formulated as follows:
\begin{equation}
\begin{split}
    {p}_{\theta}(\mathbf{y}|\mathbf{g}, \mathbf{z_1}, \mathbf{z_2}) 
    & \approx {p}_{\theta}\left(\mathbf{y} \left |\, \mathbf{g}, \frac{1}{\textit{L}} \sum_{i=1}^\textit{L} \mathbf{z}_1^i\right., \frac{1}{\textit{M}} \sum_{j=1}^\textit{M} \mathbf{z}_2^j\right)\\
    &={p}_{\theta}(\mathbf{y}|\mathbf{g}, \mathbf{\bar{z}_1}, \mathbf{\bar{z}_2}).
\end{split}
\end{equation}

The approximation error (\textit{i.e.}, \textit{Jensen gap} \cite{abramovich2016some}) can be well bounded for most functions in calculating ${p}_{\theta}(\mathbf{y}|\mathbf{g}, \mathbf{\bar{z}_1}, \mathbf{\bar{z}_2})$\cite{gao2017bounds}. Thereafter, we use a GNN to incorporate $\mathbf{g}$ with $\mathbf{\bar{z}_1}$ and $\mathbf{\bar{z}_2}$ to obtain the final embeddings:
\begin{equation}
    \mathbf{h} = \operatorname{GNN}_\emph{y}\left(\emph{A}_{\emph{g}}, [\mathbf{\bar{z}_1} \| \mathbf{\bar{z}_2}]\right),
    \label{final}
\end{equation}
where $\emph{A}_{\emph{g}}$ is the adjacency matrix of $\emph{G}_\emph{s}$. Finally, we model ${p}_{\theta}(\mathbf{y}|\mathbf{g}, \mathbf{z_1}, \mathbf{z_2})$ with an MLP that utilizes $\mathbf{h}$ to produce a probability distribution over multiple node classes.


\noindent\textbf{Instantiating ${p}_{\theta}(\mathbf{g} | \mathbf{e_1}, \mathbf{e_2})$ with Latent Space Model:} We specify ${p}_{\theta}(\mathbf{g} | \mathbf{e_1}, \mathbf{e_2})$ by modelling the distribution of edges in $\mathbf{g}$ under the condition of $\mathbf{e_1}$ and $\mathbf{e_2}$. Let $\emph{a}_\emph{ij} \in \mathcal{V} \times \mathcal{V}$ represent the binary edge random variable between node $i$ and node $j$. $\emph{a}_\emph{ij}$ = $1$ indicates the edge between $i$ and $j$ exists and $0$ otherwise. A general assumption used by various generative models for graphs is that the edges are conditionally independent \cite{ma2019flexible}. Therefore, ${p}_{\theta}(\mathbf{g} | \mathbf{e_1}, \mathbf{e_2})$ can be factorized as follows:
\begin{equation}
    p_{\theta}(\mathbf{g} | \mathbf{e_1}, \mathbf{e_2})=\prod_{\emph{i,j}} p_{\theta}\left(\emph{a}_\emph{ij} | \mathbf{e_1}, \mathbf{e_2}\right).
\end{equation}

Next, we adopt the latent space model (LSM) \cite{hoff2002latent}, which is a widely used generative network model in the network science literature. LSM assumes that the nodes lie in a latent space and the probability of $\emph{a}_\emph{ij}$ only depends on the representations of nodes $i$ and $j$, \textit{i.e.}, $p_{\theta_2}(\emph{a}_\emph{ij}|\mathbf{e_1},\mathbf{e_2})$ = $p_{\theta_2}(\emph{a}_\emph{ij}|{\mathbf{e}^i_\mathbf{1}},{\mathbf{e}^j_\mathbf{1}}, {\mathbf{e}^i_\mathbf{2}}, {\mathbf{e}^j_\mathbf{2}})$. We assume it follows a logistic regression model:
\begin{equation}
\begin{split}
    \quad&\quad p_{\theta}(\emph{a}_\emph{ij}=1 | {\mathbf{e}^i_\mathbf{1}},{\mathbf{e}^j_\mathbf{1}}, {\mathbf{e}^i_\mathbf{2}}, {\mathbf{e}^j_\mathbf{2}})\\
    &=\delta\left(\boldsymbol{w}^\mathsf{T}\left[\mathbf{U}(\mathbf{x}_i), \mathbf{U}(\mathbf{x}_j), \left[\mathbf{e}^\emph{c-i}_\mathbf{2}\|\mathbf{e}^\emph{u-i}_\mathbf{2}\right], [\mathbf{e}^\emph{c-j}_\mathbf{2}\|\mathbf{e}^\emph{u-j}_\mathbf{2}]\right] \right),
\end{split}
\end{equation}
where $\delta(\cdot)$ is the sigmoid function; $\boldsymbol{w}$ are the learnable parameters of the logistic regression model; $\mathbf{U}(\cdot)$ is a linear transformation function. $\mathbf{e}^\emph{c-i}_\mathbf{2}$ and $\mathbf{e}^\emph{u-i}_\mathbf{2}$ denote the embeddings of node $i$ sampled from ${q}_\phi(\mathbf{e_2^c}|\mathbf{y}, \mathbf{g}, \mathbf{e_1})$ and ${q}_\phi(\mathbf{e_2^u}|\mathbf{y}, \mathbf{g}, \mathbf{e_1})$, respectively. We concatenate the transformed features and node embeddings as the input of the logistic regression model. 

\subsection{Meta-Learning}
\begin{figure}[t]
\setlength{\abovecaptionskip}{0cm} 
\setlength{\belowcaptionskip}{0cm} 
\centering
\scalebox{0.256}{\includegraphics{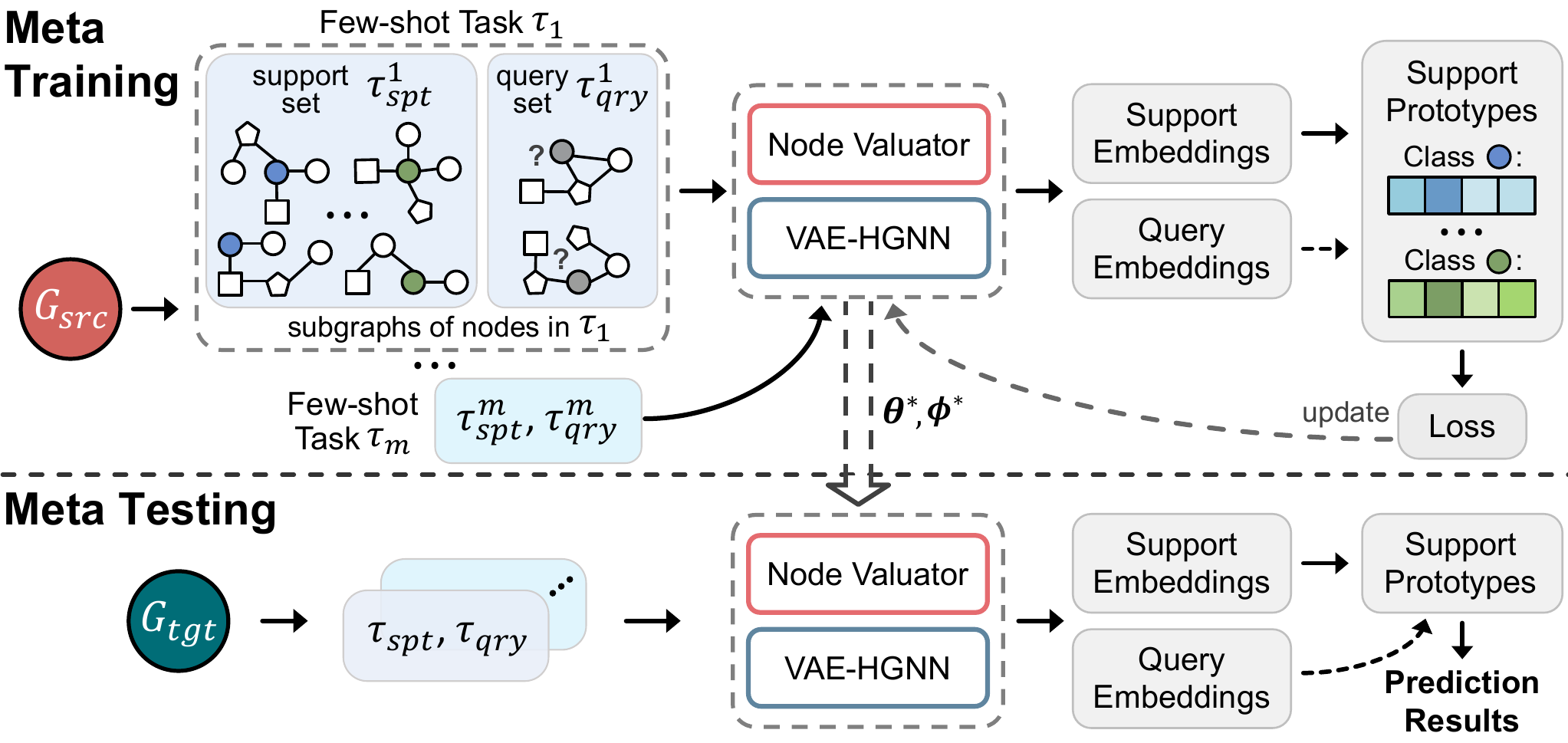}}
\caption{Meta-learning Framework.}
\label{meta}
\end{figure}

Fig. \ref{meta} illustrates an overview of our proposed meta-learning framework, which consists of two steps: meta-training and meta-testing. In meta-training, we utilize few-shot tasks on the source HG to train VAE-HGNN, where a novel \emph{node valuator} module is proposed to determine the informativeness of each labeled node. Then, we adopt a prototypical network to transfer generalized knowledge from the source HG to the target HG. During meta-testing, the node valuator is adapted to the target HG to evaluate the importance of few-labeled nodes in OOD environments, thereby facilitating the generation of robust class representations for prediction.

\noindent\textbf{Node Valuator.} Evaluating the importance of few-labeled samples is a prevalent strategy in existing graph few-shot learning methods \cite{ding2023cross,ding2020graph}. The general evaluation approach prioritizes nodes with notable centrality and popularity. However, this approach may not be appropriate for OOD environments, where the source HG and target HG might have distinct node degree distributions.


To address this issue, we adopt a causal view to evaluate the importance of labeled nodes in OOD environments. We focus on evaluating the richness of node features that remain robust against distribution shifts and determine whether these features are distinctly identifiable. Based on our proposed SCM, we consider the following two aspects: (1) Since $\emph{Z}_2$ is unaffected by distribution shifts and directly influences $\emph{Y}$, we investigate node features associated with $\emph{Z}_2$. (2) As $\emph{Z}_1$ is influenced by changes in $\emph{E}_1$ and $\emph{G}_\emph{s}$, to effectively learn the features of $\emph{Z}_2$ without interference from the features of $\emph{Z}_1$, node features related to $\emph{Z}_2$ should be distinguishable from those related to $\emph{Z}_1$. Specifically, we evaluate the node importance from the richness of its structural-level and node-level features:

\noindent \textbf{1. Structure-level features:} 
Since the semantic features of $\emph{Z}_1$ and $\emph{Z}_2$ are extracted from specific graph structures, having distinctive features related to $\emph{Z}_2$ implies that the structural differences between $\emph{Z}_2$ and $\emph{Z}_1$ should be maximized. Furthermore, the graph structure behind $\emph{Z}_2$ should exhibit a certain degree of connectivity, as insufficient connectivity may result in a structure that lacks meaningful semantics.


\noindent \textbf{2. Node-level features:} The node features in $\emph{Z}_1$ and $\emph{Z}_2$ should also be distinctly different. However, directly comparing the representations of the same node in $\emph{Z}_1$ and $\emph{Z}_2$ can be influenced by the node's inherent characteristics, which are present in both $\emph{Z}_1$ and $\emph{Z}_2$. Thus, we propose incorporating the features of neighbor nodes as a reference for effective comparison.

For the richness of structure-level features, we compare the similarity between $\emph{A}_{\emph{z}_1}^\emph{sum}$ and $\emph{A}_{\emph{z}_2}$, and assess the connectivity of $\emph{A}_{\emph{z}_2}$. Here, $\emph{A}_{\emph{z}_1}^\emph{sum}$ = $\frac{1}{l} \sum_{i=1}^l \emph{A}_{\emph{z}_1}^l$ is derived from the Multi-layer GNN module as described in Eq. (\ref{hz1}), while $\emph{A}_{\emph{z}_2}$ is obtained from the graph learner module as outlined in Eq. (\ref{z2}). The richness score of the structure-level features is defined as:
\begin{equation}
\operatorname{rsl}(v_{t}) = \cos\left(\emph{A}_{\emph{z}_1}^\emph{sum}, \emph{A}_{\emph{z}_2}\right) + \log\left(\frac{\sum_{i=1}^{|\mathcal{V}_\emph{s}|}\operatorname{deg_{\emph{z}_2}}(v_i)}{|\mathcal{V}_\emph{s}|} + \epsilon\right),
\end{equation}
where $\epsilon$ is a small constant, $\operatorname{deg_{\emph{z}_2}}(v_i)$ is the in-degree of node $v_i$ in $\emph{A}_{\emph{z}_2}$, and $|\mathcal{V}_\emph{s}|$ is the number of nodes in the subgraph of $v_t$ (\emph{i.e.}, $\emph{G}_\emph{s}$). For node-level features, we first compute the richness score for each neighbor $v_\emph{n}$ of $v_t$:
\begin{equation}
\operatorname{sn}({v_t}, {v_\emph{n}}) = \tanh\left(\mathbf{W}_\emph{s}\cdot \left[\mathbf{h}_{v_t}^{\emph{z}_1}\|\mathbf{h}_{v_\emph{n}}^{\emph{z}_2}\right]\right),
\end{equation}
where $\mathbf{h}_{v_\emph{n}}^{\emph{z}_2}$ represents the embedding of $v_\emph{n}$ sampled from ${p}_{\theta}(\mathbf{z_2} |\mathbf{e_2})$, and $\mathbf{h}_{v_t}^{\emph{z}_1}$ represents the embedding of $v_t$ sampled from ${p}_{\theta}(\mathbf{z_1} | \mathbf{g},\mathbf{e_1})$. $\mathbf{W}_\emph{s}$ is a learnable weight matrix. Then, we adopt an attention mechanism to compute the importance of the neighbor $v_\emph{n}$ and compute the node-level richness score:
\begin{gather}
\gamma_{t \emph{n}}=\frac{\exp \left(\operatorname{LeakyReLU}\left(\mathbf{a}_\emph{s}^{\mathsf{T}}\left[\mathbf{h}_{v_t}^{\emph{z}_2} \| \mathbf{h}_{v_\emph{n}}^{\emph{z}_2}\right]\right)\right)}{\sum_{v_j \in \textbf{Nei}_t} \exp \left(\operatorname{LeakyReLU}\left(\mathbf{a}_\emph{s}^{\mathsf{T}}\left[\mathbf{h}_{v_j}^{\emph{z}_2} \| \mathbf{h}_{v_\emph{n}}^{\emph{z}_2}\right]\right)\right)},
\\
\operatorname{rnl}(v_t)=\sum\nolimits_{v_i \in \textbf{Nei}_t}\gamma_{t i}\cdot\operatorname{sn}({v_{t}}, {v_i}),
\end{gather}
where $\textbf{Nei}_t$ is the set of neighbors of $v_t$ in $\emph{A}_{\emph{z}_2}$, and $\mathbf{a}_\emph{s}$ is the trainable attention parameter. Finally, the richness score of a labeled node can be computed as $\operatorname{rs}(v_{t})= \operatorname{rsl}(v_{t})+\operatorname{rnl}(v_{t})$.

\noindent\textbf{Prototypical Network.}
We follow the idea of the prototypical network \cite{snell2017prototypical} and calculate the weighted average of $\emph{K}$-shot embedded nodes within class ${c}$ to derive the class prototype:
\begin{equation}\label{proto_fun}
\mathbf{proto}_{c} = \sum\nolimits^\emph{K}_{{i}=1}\operatorname{sc}_{i}\cdot \mathbf{h}_{v_i},
\end{equation}
where $\operatorname{sc}_{i}$ = $\operatorname{rs}(v_{i})/\sum\nolimits^\emph{K}_{{j}=1}\operatorname{rs}(v_{j})$ is the normalized score, $\mathbf{h}_{v_i}$ is the final embedding of node $v_{i}$ output by VAE-HGNN. The node valuator module is trained through meta-training and is used to compute the class prototype in meta-testing.

\noindent\textbf{Loss Function.} 
During meta-training, the prototype for each class is computed using the nodes in the support set $\tau^\emph{spt}$. To determine the class of a query node $u$ in the query set $\tau^\emph{qry}$, we calculate the probability for each class based on the distance between the node embedding $\mathbf{h}_{u}$ and each prototype:
\begin{equation}\label{class_vec}
\operatorname{prob}({c}|{u})=\frac{\exp \left(-\operatorname{d}(\mathbf{h}_{u},\mathbf{proto}_{c})\right)}{\sum_{c'} \exp \left(-\operatorname{d}(\mathbf{h}_{u},\mathbf{proto}_{c'})\right)},
\end{equation}
where $\operatorname{d}(\cdot)$ is a distance metric function and we adopt squared Euclidean distance \cite{snell2017prototypical}. Under the episodic training framework, the objective of each meta-training task is to minimize the classification loss between the predictions of the query set and the ground truth. The training loss for a single task $\tau$ is the average negative log-likelihood probability of assigning correct class labels as follows:
\begin{equation}\label{loss_task}
\mathcal{L}_\tau = -\frac{1}{\emph{N}\times \emph{K}}\sum\nolimits_{{i}=1}^{\emph{N}\times \emph{K}} \log \left(\operatorname{prob}({y}^*_{i}|{v}_{i})\right),
\end{equation}
where ${y}^*_{i}$ is the ground truth label of ${v}_{i}$. Then, by incorporating VAE-HGNN with the ELBO loss ($\mathcal{L}_\emph{ELBO}$ = $\mathcal{L}_\emph{cls}+\mathcal{L}_\emph{str}+\mathcal{L}_\emph{kl}$). $\operatorname{prob}({y}^*_{i}|{v}_{i})$ can be modeled by ${p}_{\theta}(\mathbf{y}|\mathbf{g},\mathbf{e_1},\mathbf{e_2})$ and thus $\mathcal{L}_\tau$ = $\mathcal{L}_\emph{cls}$. Finally, the total meta-training loss can be defined as:
\begin{equation}\label{loss_meta}
\mathcal{L}_\emph{meta} = \sum\nolimits_{\emph{j}=1}^{|\mathcal{T}_\emph{src}|}\mathcal{L}^j_{\emph{cls}} + \mathcal{L}^j_\emph{str}+\lambda\cdot\mathcal{L}^j_\emph{kl},
\end{equation}
where $\lambda$ is a loss coefficient hyper-parameter to restrict the regularization of the KL divergence.

\subsection{Efficiency Analysis}
Targeting the novel and complex OOD generalization in HG few-shot learning, our proposed COHF is efficient due to the following properties:

\noindent\textbf{Subgraph-based Few-shot Learning:} 
COHF extracts subgraphs around labeled nodes for learning, instead of processing the full graph structure. By incorporating causal learning, COHF further focuses on structures that have a causal relationship with the labels. This strategy ensures that both the input graph size and the learning graph structure remain small and consistent during computation, leading to stable memory consumption and improved computational efficiency.
 
\noindent\textbf{Relation-based Semantic Extraction in OOD environments:} Traditional HGNNs typically adopt multiple modules to capture a variety of heterogeneous information for learning semantics, which becomes time-consuming when processing multiple HGs with diverse heterogeneity. In contrast, COHF extracts heterogeneous information only from common and unique relations, which not only simplifies the few-shot learning framework in OOD environments, but also ensures that COHF's computational complexity remains unaffected by the complexity of the heterogeneous information.

Furthermore, we conduct experiments to compare the time and memory consumption of COHF and the baselines in the same OOD scenarios. See Appendix III for comparison results.

\section{Experiments and Analysis}
We conduct extensive experiments on seven real-world datasets to answer the following research questions: \textbf{RQ1:} How does COHF perform compared with representative and state-of-the-art methods in OOD environments? \textbf{RQ2:} How does each of the proposed VAE-HGNN module and node valuator module contribute to the overall performance? \textbf{RQ3:} How does COHF perform under different parameter settings?

\subsection{Experimental Settings}
\noindent\textbf{Datasets.} 
We adopt seven real-world datasets from three domains: (1) ACM\cite{yun2019graph} and DBLP\cite{fu2020magnn} from the academic research domain. (2) IMDB\cite{fu2020magnn}, MovieLens\cite{miller2003movielens}, and Douban\cite{wang2022survey} from the movie domain. (3) YELP-Restaurant (abbreviated as YELP-R) \cite{shi2020rhine} and YELP-Business (abbreviated as YELP-B) \cite{yang2020heterogeneous} from the business domain. These datasets are publicly available and have been widely used in studies of HGs \cite{wang2022survey}. Details of these datasets are provided in Table \ref{dataset}.

\begin{table}[]
\setlength{\abovecaptionskip}{0cm} 
\setlength{\belowcaptionskip}{0cm} 
\centering
\caption{Statistics of Datasets.}
\label{dataset}
\resizebox{79mm}{45mm}{
\begin{tabular}{c|c|c|c}
\specialrule{0.15em}{1pt}{1pt} 
\textbf{Domain}                   & \textbf{Dataset}   & \textbf{\# Nodes}                                                                                                                                           & \textbf{\# Relations}                                                                                                               \\\specialrule{0.05em}{0.7pt}{0.7pt}
\multirow{5}{*}{Academic} & ACM       & \begin{tabular}[c]{@{}c@{}}\# authors (A): 5,912\\ \# papers (P): 3,025\\ \# subjects (S): 57\end{tabular}                                                 & \begin{tabular}[c]{@{}c@{}}\# A-P: 9,936\\ \# P-S: 3,025\end{tabular}                                                       \\
\cmidrule(r){2-4}
                          & DBLP      & \begin{tabular}[c]{@{}c@{}}\# authors (A): 4,057\\ \# papers (P): 14,328\\ \# terms (T): 7,723\\ \# venues (V): 20\end{tabular}                               & \begin{tabular}[c]{@{}c@{}}\# A-P: 19,645\\ \# P-T: 85,810\\ \# P-V: 14,328\end{tabular}                                     \\\specialrule{0.05em}{0.7pt}{0.7pt}
\multirow{9}{*}{Movie}    & IMDB      & \begin{tabular}[c]{@{}c@{}}\# movies (M): 4,278 \\ \# directors (D): 2,081 \\ \# actors (A): 5,257\end{tabular}                                         & \begin{tabular}[c]{@{}c@{}}\# M-D: 4,278 \\ \# M-A: 12,828\end{tabular}                                                   \\
\cmidrule(r){2-4}
                          & MovieLens & \begin{tabular}[c]{@{}c@{}}\# movies (M): 1,682 \\ \# users (U): 943 \\ \# occupations (O): 21\\ \# ages (E): 8\end{tabular}                              & \begin{tabular}[c]{@{}c@{}}\# U-M: 100,000\\ \# U-E: 943\\ \# U-O: 943\end{tabular}                                          \\
                          \cmidrule(r){2-4}
                          & Douban    & \begin{tabular}[c]{@{}c@{}}\# movies (M): 12,677 \\ \# directors (D): 2,449 \\ \# actors (A): 6,311\\ \# users (U): 13,367\\ \# groups (G): 2,753\end{tabular} & \begin{tabular}[c]{@{}c@{}}\# M-D: 11,276 \\ \# M-A: 33,587\\ \# U-M: 355,072\\ \# U-G: 127,632\\ \# U-U: 392,519\end{tabular} \\
                          \specialrule{0.05em}{0.7pt}{0.7pt}
\multirow{5}{*}{Business} & YELP-R    & \begin{tabular}[c]{@{}c@{}}\# businesses (B): 2,614\\ \# users (U): 1,286\\ \# services (S): 2\\ \# star levels (L): 9\\ \# reservations (R): 2\end{tabular}      & \begin{tabular}[c]{@{}c@{}}\# B-U: 30,838\\ \# B-R: 2,614\\ \# B-S: 2,614\\ \# B-L: 2,614\end{tabular}                        \\\cmidrule(r){2-4}
                          & YELP-B    & \begin{tabular}[c]{@{}c@{}}\# businesses (B): 7,474\\ \# phrases (P): 74,943\\ \# star levels (L): 9\\ \# locations (O): 39\end{tabular}                       & \begin{tabular}[c]{@{}c@{}}\# B-P: 27,209,836\\ \# B-L: 7,474\\ \# B-O: 7,474\\ \# P-P: 2,654,313\end{tabular}  \\\specialrule{0.15em}{1pt}{1pt}          
\end{tabular}}
\end{table}

\noindent\textbf{Meta-learning and OOD Settings.} 
We consider the following two distinct scenarios: 

\begin{table*}[]
\setlength{\abovecaptionskip}{0cm} 
\setlength{\belowcaptionskip}{0cm} 
\centering
\caption{Node classification accuracy in inter-dataset setting.}
\label{exp_res}
\resizebox{174mm}{65.6mm}{
\setlength{\tabcolsep}{1mm}{
\begin{tabular}{ccccccccccccccccc}
\specialrule{0.15em}{2pt}{2pt} 
\multicolumn{1}{c|}{}            & \multicolumn{2}{c}{\textbf{ACM-DBLP}}           & \multicolumn{2}{c|}{\textbf{DBLP-ACM}}                               & \multicolumn{2}{c}{\textbf{IMDB-Douban}}           & \multicolumn{2}{c|}{\textbf{Douban-IMDB}}                               & \multicolumn{2}{c}{\textbf{MvLens-Douban}}           & \multicolumn{2}{c|}{\textbf{Douban-MvLens}}                               & \multicolumn{2}{c}{\textbf{YELP R-B}}           & \multicolumn{2}{c}{\textbf{YELP B-R}}           \\\specialrule{0.03em}{0.7pt}{0.7pt}
\multicolumn{1}{c|}{}            & \textbf{2-way}           & \textbf{3-way}          & \textbf{2-way}           & \multicolumn{1}{c|}{\textbf{3-way} }           & \textbf{2-way}           & \textbf{3-way}           & \textbf{2-way}           & \multicolumn{1}{c|}{\textbf{3-way} }           & \textbf{2-way}           & \textbf{3-way}            & \textbf{2-way}           & \multicolumn{1}{c|}{\textbf{3-way} }           & \textbf{2-way}           & \textbf{3-way}            & \textbf{2-way}           & \textbf{3-way}           \\\specialrule{0.03em}{0.7pt}{0.7pt}
\multicolumn{17}{c}{\cellcolor[HTML]{E5E5E5}$\textbf{\emph{1-shot}}$}    \\\specialrule{0.03em}{0.7pt}{0.7pt}
\multicolumn{1}{c|}{GAT}         & 0.7133          & 0.5489          & 0.6297          & \multicolumn{1}{c|}{0.4733}          & 0.4017          & 0.3052          & 0.5167          & \multicolumn{1}{c|}{0.3458}          & 0.4590          & 0.2907          & 0.5050          & \multicolumn{1}{c|}{0.3458}          & 0.5043          & 0.3413          & 0.5470          & 0.3544          \\
\multicolumn{1}{c|}{SGC}         & 0.7147          & 0.6057          & 0.5698          & \multicolumn{1}{c|}{0.4652}          & 0.4932          & 0.3235          & 0.5147          & \multicolumn{1}{c|}{0.3512}          & 0.4953          & 0.3269          & 0.5174          & \multicolumn{1}{c|}{0.3414}          & 0.5077          & 0.3469          & 0.5118          & 0.3625          \\
\multicolumn{1}{c|}{GIN}         & 0.6984          & 0.5106          & 0.5982          & \multicolumn{1}{c|}{0.4522}          & 0.4453          & 0.2909          & 0.5222          & \multicolumn{1}{c|}{0.3615}          & 0.4431          & 0.2591          & 0.5117          & \multicolumn{1}{c|}{0.3475}          & 0.5126          & 0.3370          & 0.5930          & 0.4368          \\\specialrule{0.03em}{0.7pt}{0.7pt}
\multicolumn{1}{c|}{HAN}         & 0.8088          & 0.6417          & 0.7453*         & \multicolumn{1}{c|}{0.6071}          & 0.4937          & 0.3240          & 0.5240          & \multicolumn{1}{c|}{0.3622}          & 0.4313          & 0.2767          & 0.5340          & \multicolumn{1}{c|}{0.3531}          & 0.5283          & 0.3651          & 0.6417          & 0.4633          \\
\multicolumn{1}{c|}{MAGNN}       & 0.8112          & 0.6713          & 0.7359          & \multicolumn{1}{c|}{0.6172*}         & 0.4967          & 0.3252          & 0.5355          & \multicolumn{1}{c|}{0.3659}          & 0.5150          & 0.3469*         & 0.5411          & \multicolumn{1}{c|}{0.3804}          & 0.5278          & 0.3607          & 0.6244          & 0.4674          \\\specialrule{0.03em}{0.7pt}{0.7pt}
\multicolumn{1}{c|}{UDA-GCN}         & 0.5347          & 0.3456          & 0.5167         & \multicolumn{1}{c|}{0.3440}          & 0.5007          & 0.3418          & 0.5060          & \multicolumn{1}{c|}{0.3507}          & 0.4954          & 0.3224          & 0.5001          & \multicolumn{1}{c|}{0.3332}          & 0.5144          & 0.3398          & 0.5223          & 0.3316          \\
\multicolumn{1}{c|}{MuSDAC}       & 0.5160          & 0.3578          & 0.5233          & \multicolumn{1}{c|}{0.3556}         & 0.5047          & 0.3476          & 0.5127          & \multicolumn{1}{c|}{0.3453}          & 0.5093          & 0.3413         & 0.5067          & \multicolumn{1}{c|}{0.3409}          & 0.5133          & 0.3418          & 0.5273          & 0.3578          \\\specialrule{0.03em}{0.7pt}{0.7pt}
\multicolumn{1}{c|}{CAL}         & 0.5023          & 0.3288          & 0.5490          & \multicolumn{1}{c|}{0.3296}          & 0.4898          & 0.3219          & 0.5094          & \multicolumn{1}{c|}{0.3370}          & 0.4950          & 0.3367          & 0.4913          & \multicolumn{1}{c|}{0.3352}          & 0.5063          & 0.3339          & 0.5060          & 0.3412          \\
\multicolumn{1}{c|}{DIR}         & 0.5939          & 0.4880          & 0.6359          & \multicolumn{1}{c|}{0.4943}          & 0.4313          & 0.2710          & 0.5117          & \multicolumn{1}{c|}{0.3477}          & 0.4283          & 0.2632          & 0.4871          & \multicolumn{1}{c|}{0.3229}          & 0.5027          & 0.3345          & 0.5722          & 0.4284          \\
\multicolumn{1}{c|}{WSGNN}       & 0.5300          & 0.3489          & 0.5037          & \multicolumn{1}{c|}{0.3347}          & 0.5003          & 0.3321          & 0.4828          & \multicolumn{1}{c|}{0.3240}          & 0.5017          & 0.3345          & 0.4767          & \multicolumn{1}{c|}{0.3408}          & 0.5117          & 0.3186          & 0.4834          & 0.3267          \\\specialrule{0.03em}{0.7pt}{0.7pt}
\multicolumn{1}{c|}{MAML}        & 0.6339          & 0.4551          & 0.5740          & \multicolumn{1}{c|}{0.3896}          & 0.4991          & 0.3285          & 0.4971          & \multicolumn{1}{c|}{0.3345}          & 0.4975          & 0.3332          & 0.5008          & \multicolumn{1}{c|}{0.3384}          & 0.4976          & 0.3317          & 0.5109          & 0.3436          \\
\multicolumn{1}{c|}{ProtoNet}    & 0.5580          & 0.3949          & 0.5636          & \multicolumn{1}{c|}{0.3939}          & 0.4838          & 0.3119          & 0.5068          & \multicolumn{1}{c|}{0.3436}          & 0.4758          & 0.2986          & 0.5157          & \multicolumn{1}{c|}{0.3473}          & 0.4990          & 0.3338          & 0.5026          & 0.3439          \\\specialrule{0.03em}{0.7pt}{0.7pt}
\multicolumn{1}{c|}{Meta-GNN}    & 0.7099          & 0.5915          & 0.6509          & \multicolumn{1}{c|}{0.5312}          & 0.4844          & 0.3261          & 0.5167          & \multicolumn{1}{c|}{0.3448}          & 0.4924          & 0.3342          & 0.5227          & \multicolumn{1}{c|}{0.3523}          & 0.5002          & 0.3324          & 0.5316          & 0.3536          \\
\multicolumn{1}{c|}{GPN}         & 0.6122          & 0.4518          & 0.7099          & \multicolumn{1}{c|}{0.5355}          & 0.4501          & 0.2920          & 0.5032          & \multicolumn{1}{c|}{0.3389}          & 0.5004          & 0.3337          & 0.4928          & \multicolumn{1}{c|}{0.3335}          & 0.5001          & 0.3334          & 0.5127          & 0.4016          \\
\multicolumn{1}{c|}{G-Meta}      & 0.6879          & 0.5377          & 0.5744          & \multicolumn{1}{c|}{0.4007}          & 0.4907          & 0.3404*         & 0.5370          & \multicolumn{1}{c|}{0.3273}          & 0.4795          & 0.3301          & 0.5194          & \multicolumn{1}{c|}{0.3929*}         & 0.5145          & 0.3665          & 0.5291          & 0.3768          \\\specialrule{0.03em}{0.7pt}{0.7pt}
\multicolumn{1}{c|}{CGFL}        & 0.8240*         & 0.7211*         & 0.7413          & \multicolumn{1}{c|}{0.6049}          & 0.5150*         & 0.3080          & 0.5374*         & \multicolumn{1}{c|}{0.3714*}         & 0.5290*         & 0.3153          & 0.5467          & \multicolumn{1}{c|}{0.3701}          & 0.5358*         & 0.3769*         & 0.6509*         & 0.4797*         \\
\multicolumn{1}{c|}{HG-Meta}     & 0.7940          & 0.7031          & 0.7337          & \multicolumn{1}{c|}{0.5631}          & 0.5038          & 0.3176          & 0.5320          & \multicolumn{1}{c|}{0.3653}          & 0.4610          & 0.2944          & 0.5514*         & \multicolumn{1}{c|}{0.3840}          & 0.5216          & 0.3489          & 0.6316          & 0.4609          \\\specialrule{0.03em}{0.7pt}{0.7pt}
\multicolumn{1}{c|}{COHF}        & $\textbf{0.8416}^1$ & \textbf{0.7375} & \textbf{0.7753} & \multicolumn{1}{c|}{\textbf{0.6407}} & \textbf{0.5316} & \textbf{0.3611} & \textbf{0.5680} & \multicolumn{1}{c|}{\textbf{0.3944}} & \textbf{0.5584} & \textbf{0.3944} & \textbf{0.5916} & \multicolumn{1}{c|}{\textbf{0.4067}} & \textbf{0.5709} & \textbf{0.4027} & \textbf{0.6722} & \textbf{0.4896} \\
\multicolumn{1}{c|}{Improvement$^2$} & \underline{2.14\%}          & \underline{2.27\%}          & \underline{4.03\%}          & \multicolumn{1}{c|}{\underline{3.81\%}}          & \underline{3.22\%}          & \underline{6.08\%}          & \underline{5.69\%}          & \multicolumn{1}{c|}{\underline{6.19\%}}          & \underline{5.56\%}          & \underline{13.69\%}          & \underline{7.29\%}          & \multicolumn{1}{c|}{\underline{3.51\%}}          & \underline{6.55\%}          & \underline{6.85\%}          & \underline{3.27\%}          & \underline{2.06\%}          \\\specialrule{0.05em}{1pt}{1pt}
\multicolumn{17}{c}{\cellcolor[HTML]{E5E5E5}$\textbf{\emph{3-shot}}$}    \\\specialrule{0.03em}{0.7pt}{0.7pt}
\multicolumn{1}{c|}{GAT}         & 0.8162          & 0.6733          & 0.6760          & \multicolumn{1}{c|}{0.5549}          & 0.4067          & 0.3122          & 0.5429          & \multicolumn{1}{c|}{0.3880}          & 0.4416          & 0.2622          & 0.5355          & \multicolumn{1}{c|}{0.4014}          & 0.5189          & 0.3469          & 0.5509          & 0.4087          \\
\multicolumn{1}{c|}{SGC}         & 0.8107          & 0.6788          & 0.6944          & \multicolumn{1}{c|}{0.5577}          & 0.4878          & 0.3104          & 0.5461          & \multicolumn{1}{c|}{0.3707}          & 0.4790          & 0.3235          & 0.5065          & \multicolumn{1}{c|}{0.4050}          & 0.5031          & 0.3478          & 0.5416          & 0.4378          \\
\multicolumn{1}{c|}{GIN}         & 0.7694          & 0.6204          & 0.6787          & \multicolumn{1}{c|}{0.5542}          & 0.3909          & 0.2975          & 0.5457          & \multicolumn{1}{c|}{0.3858}          & 0.4764          & 0.2428          & 0.5429          & \multicolumn{1}{c|}{0.4376}          & 0.5093          & 0.3619          & 0.6662          & 0.5118          \\\specialrule{0.03em}{0.7pt}{0.7pt}
\multicolumn{1}{c|}{HAN}         & 0.8795          & 0.7532          & 0.8233          & \multicolumn{1}{c|}{0.6052}          & 0.4923          & 0.3091          & 0.5583          & \multicolumn{1}{c|}{0.3962*}         & 0.3770          & 0.2824          & 0.5631          & \multicolumn{1}{c|}{0.3932}          & 0.5267          & 0.3667          & 0.6811          & 0.5378          \\
\multicolumn{1}{c|}{MAGNN}       & 0.8912          & 0.7641          & 0.8325*         & \multicolumn{1}{c|}{0.5227}          & 0.5075*         & 0.3445          & 0.5461          & \multicolumn{1}{c|}{0.3937}          & 0.5204          & 0.3441          & 0.5849          & \multicolumn{1}{c|}{0.4522*}         & 0.5255          & 0.3533          & 0.6930          & 0.4636          \\\specialrule{0.03em}{0.7pt}{0.7pt}
\multicolumn{1}{c|}{UDA-GCN}         & 0.5153          & 0.3498          & 0.5520          & \multicolumn{1}{c|}{0.3480}          & 0.5047          & 0.3442          & 0.5120          & \multicolumn{1}{c|}{0.3453}         & 0.5004          & 0.3222          & 0.5001          & \multicolumn{1}{c|}{0.3331}          & 0.5004          & 0.3341          & 0.5322          & 0.3517          \\
\multicolumn{1}{c|}{MuSDAC}       & 0.5224          & 0.3664          & 0.5455         & \multicolumn{1}{c|}{0.3882}          & 0.5067         & 0.3431          & 0.5147          & \multicolumn{1}{c|}{0.3422}          & 0.4987          & 0.3258          & 0.5147          & \multicolumn{1}{c|}{0.3489}         & 0.5180          & 0.3413          & 0.5573          & 0.3661          \\\specialrule{0.03em}{0.7pt}{0.7pt}
\multicolumn{1}{c|}{CAL}         & 0.5193          & 0.3382          & 0.4974          & \multicolumn{1}{c|}{0.3303}          & 0.4917          & 0.3229          & 0.5183          & \multicolumn{1}{c|}{0.3352}          & 0.5053          & 0.3393          & 0.5307          & \multicolumn{1}{c|}{0.3370}          & 0.5106          & 0.3313          & 0.5130          & 0.3457          \\
\multicolumn{1}{c|}{DIR}         & 0.6897          & 0.5600          & 0.6760          & \multicolumn{1}{c|}{0.5371}          & 0.4333          & 0.2880          & 0.5339          & \multicolumn{1}{c|}{0.3658}          & 0.4343          & 0.2932          & 0.5060          & \multicolumn{1}{c|}{0.3400}          & 0.5157          & 0.3547          & 0.6323          & 0.4704          \\
\multicolumn{1}{c|}{WSGNN}       & 0.4518          & 0.3378          & 0.5183          & \multicolumn{1}{c|}{0.3412}          & 0.5011          & 0.3444          & 0.5122          & \multicolumn{1}{c|}{0.3341}          & 0.3586          & 0.3370          & 0.5011          & \multicolumn{1}{c|}{0.3526}          & 0.4822          & 0.3489          & 0.4878          & 0.3378          \\\specialrule{0.03em}{0.7pt}{0.7pt}
\multicolumn{1}{c|}{MAML}        & 0.7288          & 0.5409          & 0.6248          & \multicolumn{1}{c|}{0.4968}          & 0.4962          & 0.3345          & 0.5084          & \multicolumn{1}{c|}{0.3347}          & 0.5014          & 0.3369          & 0.4986          & \multicolumn{1}{c|}{0.3409}          & 0.5098          & 0.3413          & 0.5052          & 0.3572          \\
\multicolumn{1}{c|}{ProtoNet}    & 0.5634          & 0.4175          & 0.5438          & \multicolumn{1}{c|}{0.4285}          & 0.4992          & 0.3325          & 0.5036          & \multicolumn{1}{c|}{0.3397}          & 0.4988          & 0.3307          & 0.5174          & \multicolumn{1}{c|}{0.3495}          & 0.5066          & 0.3392          & 0.5062          & 0.3544          \\\specialrule{0.03em}{0.7pt}{0.7pt}
\multicolumn{1}{c|}{Meta-GNN}    & 0.8328          & 0.6939          & 0.7429          & \multicolumn{1}{c|}{0.6126}          & 0.4953          & 0.3341          & 0.5443          & \multicolumn{1}{c|}{0.3708}          & 0.4998          & 0.3435          & 0.5502          & \multicolumn{1}{c|}{0.3759}          & 0.4995          & 0.3334          & 0.5498          & 0.3625          \\
\multicolumn{1}{c|}{GPN}         & 0.7631          & 0.5636          & 0.6726          & \multicolumn{1}{c|}{0.4903}          & 0.4542          & 0.3015          & 0.5154          & \multicolumn{1}{c|}{0.3461}          & 0.4915          & 0.3415          & 0.4703          & \multicolumn{1}{c|}{0.3071}          & 0.5120           & 0.3576          & 0.5710           & 0.4569          \\
\multicolumn{1}{c|}{G-Meta}      & 0.7324          & 0.5114          & 0.6958          & \multicolumn{1}{c|}{0.6027}          & 0.4965          & 0.3569*         & 0.4919          & \multicolumn{1}{c|}{0.3411}          & 0.5056          & 0.3433          & 0.6056*         & \multicolumn{1}{c|}{0.3778}          & 0.5247          & 0.3912*         & 0.6833          & 0.3222          \\\specialrule{0.03em}{0.7pt}{0.7pt}
\multicolumn{1}{c|}{CGFL}        & 0.9241*         & 0.8178*         & 0.8207          & \multicolumn{1}{c|}{0.6368}          & 0.4956          & 0.3444          & 0.5643*         & \multicolumn{1}{c|}{0.3896}          & 0.5398*         & 0.3571          & 0.5823          & \multicolumn{1}{c|}{0.4116}          & 0.5404*         & 0.3815          & 0.6947*         & 0.5429*         \\
\multicolumn{1}{c|}{HG-Meta}     & 0.9062          & 0.8136          & 0.7882          & \multicolumn{1}{c|}{0.6491*}         & 0.4672          & 0.2985          & 0.5564          & \multicolumn{1}{c|}{0.3911}          & 0.4766          & 0.3628*         & 0.5988          & \multicolumn{1}{c|}{0.4424}          & 0.5254          & 0.3575          & 0.5858          & 0.4224          \\\specialrule{0.03em}{0.7pt}{0.7pt}
\multicolumn{1}{c|}{COHF}        & \textbf{0.9316} & \textbf{0.8263} & \textbf{0.8722} & \multicolumn{1}{c|}{\textbf{0.6778}} & \textbf{0.5512} & \textbf{0.3889} & \textbf{0.6083} & \multicolumn{1}{c|}{\textbf{0.4319}} & \textbf{0.5722} & \textbf{0.4056} & \textbf{0.6408} & \multicolumn{1}{c|}{\textbf{0.4612}} & \textbf{0.5966} & \textbf{0.4194} & \textbf{0.7307} & \textbf{0.5611} \\
\multicolumn{1}{c|}{Improvement$^2$} & \underline{0.81\%}          & \underline{1.04\%}          & \underline{4.77\%}          & \multicolumn{1}{c|}{\underline{4.42\%}}          & \underline{8.61\%}          & \underline{8.97\%}          & \underline{7.80\%}          & \multicolumn{1}{c|}{\underline{9.01\%}}          & \underline{6.00\%}          & \underline{11.80\%}          & \underline{5.81\%}          & \multicolumn{1}{c|}{\underline{1.99\%}}          & \underline{10.40\%}          & \underline{7.21\%}          & \underline{5.18\%}          & \underline{3.35\%}         \\\specialrule{0.15em}{1pt}{1pt} 
\end{tabular}}}
\scriptsize{\leftline{* Result of the best-performing baseline. $^1$ Results of the best-performing method are in bold. $^2$ Improvements of COHF over the best-performing baseline are underlined.}}
\end{table*}

\begin{table*}[t]
\setlength{\abovecaptionskip}{0cm} 
\setlength{\belowcaptionskip}{0cm} 
\caption{Node classification F1-score in 2-way 3-shot setting.}
\label{exp-f1}
\centering
\resizebox{181mm}{11.5mm}{
\setlength{\tabcolsep}{0.4mm}{
\begin{tabular}{c|ccc|cc|cc|ccc|cc|ccc|cc|cc}
\specialrule{0.15em}{1.5pt}{1.5pt}
                         & \textbf{GAT} & \textbf{SGC} & \textbf{GIN} & \textbf{HAN} & \textbf{MAGNN} & \textbf{UDA-GCN} & \textbf{MuSDAC} & \textbf{CAL} & \textbf{DIR} & \textbf{WSGNN} & \textbf{MAML} & \textbf{ProtoNet} & \textbf{Meta-GNN} & \textbf{GPN} & \textbf{G-Meta} & \textbf{CGFL} & \textbf{HG-Meta} & \textbf{COHF}   & \textbf{Improv.$^2$} \\\specialrule{0.05em}{1pt}{1pt}
\textbf{ACM-DBLP}        & 0.8114       & 0.7973       & 0.7639       & 0.8648       & 0.8896      & 0.4889  & 0.5022    & 0.5193       & 0.6419       & 0.4540         & 0.6822        & 0.4736            & 0.8241            & 0.7520       & 0.6926          & 0.9136*       & 0.9041           & \textbf{0.9289$^1$} & \underline{1.67\%}               \\
\textbf{Douban-MvLens}   & 0.5046       & 0.5056       & 0.5188       & 0.5486       & 0.5588     & 0.4532         & 0.5040     & 0.5285       & 0.4269       & 0.4710         & 0.4714        & 0.4452            & 0.5389            & 0.4208       & 0.6004*         & 0.5802        & 0.5807           & \textbf{0.6366} & \underline{6.03\%}               \\
\textbf{YELP B-R}        & 0.5269       & 0.5478       & 0.6172       & 0.6809       & 0.6879       & 0.5114         & 0.5367  & 0.4883       & 0.6127       & 0.4856         & 0.4887        & 0.4476            & 0.5456            & 0.5485       & 0.6812          & 0.6973*       & 0.5692           & \textbf{0.7251} & \underline{3.99\%}               \\\specialrule{0.05em}{1pt}{1pt}
\textbf{MvLens (I.I.D.)} & 0.5939       & 0.6012       & 0.6083       & 0.6501*      & 0.6345    & 0.5122         & 0.5113      & 0.5110       & 0.5362       & 0.5132         & 0.5090        & 0.4324            & 0.6105            & 0.5949       & 0.6447          & 0.6479        & 0.6426           & \textbf{0.6822} & \underline{4.94\%}               \\
\textbf{MvLens (OOD)}    & 0.5468       & 0.5134       & 0.5467       & 0.5813       & 0.5881     & 0.4932         & 0.4927    & 0.4973       & 0.4495       & 0.4697         & 0.4899        & 0.4138            & 0.5559            & 0.5353       & 0.6055          & 0.6183*       & 0.5887           & \textbf{0.6645} & \underline{7.47\%}               \\
\textbf{Drop}            & 7.93\%       & 14.60\%      & 10.13\%      & 10.58\%      & 7.31\%    & 3.71\%         & 3.64\%     & 2.68\%       & 16.17\%      & 8.48\%         & 3.75\%        & 4.30\%            & 8.94\%            & 10.02\%      & 6.08\%          & 4.57\%        & 8.39\%           & 2.59\%          &  \\\specialrule{0.15em}{1.5pt}{1.5pt}                
\end{tabular}}}
\scriptsize{\leftline{* Result of the best-performing baseline. $^1$ Results of the best-performing method are in bold. $^2$ Improvements of COHF over the best-performing baseline are underlined.}}
\end{table*}

\begin{table}[t]
\setlength{\abovecaptionskip}{0cm} 
\setlength{\belowcaptionskip}{0cm} 
\centering
\caption{Node classification accuracy in intra-dataset setting.}
\label{exp_acc_inter}
\resizebox{86mm}{60mm}{
\setlength{\tabcolsep}{0.5mm}{
\begin{tabular}{cccccccccc}
\specialrule{0.15em}{2pt}{2pt} 
\multicolumn{1}{c|}{}            & \multicolumn{3}{c|}{\textbf{Douban}}                                      & \multicolumn{3}{c|}{\textbf{MovieLens}}                                   & \multicolumn{3}{c}{\textbf{YELP-B}}                  \\
\multicolumn{1}{c|}{}            & \textbf{I.I.D.}           & \textbf{OOD}             & \multicolumn{1}{c|}{\textbf{Drop}}    & \textbf{I.I.D.}           & \textbf{OOD}             & \multicolumn{1}{c|}{\textbf{Drop}}    & \textbf{I.I.D.}           & \textbf{OOD}             & \textbf{Drop}    \\\specialrule{0.05em}{1pt}{1pt}
\multicolumn{10}{c}{\cellcolor[HTML]{E5E5E5}$\textbf{\emph{2-way 1-shot}}$}    \\\specialrule{0.05em}{1pt}{1pt}
\multicolumn{1}{c|}{GAT}         & 0.5107          & 0.4560          & \multicolumn{1}{c|}{10.71\%} & 0.5477          & 0.5137          & \multicolumn{1}{c|}{6.21\%}  & 0.5107          & 0.4933          & 3.41\%  \\
\multicolumn{1}{c|}{SGC}         & 0.5093          & 0.4983          & \multicolumn{1}{c|}{2.16\%}  & 0.5563          & 0.5183          & \multicolumn{1}{c|}{6.83\%}  & 0.5280          & 0.5152          & 2.42\%  \\
\multicolumn{1}{c|}{GIN}         & 0.5153          & 0.4813          & \multicolumn{1}{c|}{6.60\%}  & 0.5743          & 0.5387          & \multicolumn{1}{c|}{6.20\%}  & 0.5283          & 0.5094          & 3.58\%  \\\specialrule{0.05em}{1pt}{1pt}
\multicolumn{1}{c|}{HAN}         & 0.5473          & 0.5233*         & \multicolumn{1}{c|}{4.39\%}  & 0.5833          & 0.5567          & \multicolumn{1}{c|}{4.56\%}  & 0.5517          & 0.5217          & 5.44\%  \\
\multicolumn{1}{c|}{MAGNN}       & 0.5522          & 0.5156          & \multicolumn{1}{c|}{6.63\%}  & 0.6022          & 0.5633          & \multicolumn{1}{c|}{6.46\%}  & 0.5544          & 0.5133          & 7.41\%  \\\specialrule{0.05em}{1pt}{1pt}
\multicolumn{1}{c|}{UDA-GCN}         & 0.5100          & 0.5042         & \multicolumn{1}{c|}{1.14\%}  & 0.5033          & 0.5001          & \multicolumn{1}{c|}{0.64\%}  & 0.5140          & 0.5101          & 0.76\%  \\
\multicolumn{1}{c|}{MuSDAC}       & 0.5153          & 0.5087          & \multicolumn{1}{c|}{1.28\%}  & 0.5087          & 0.5060          & \multicolumn{1}{c|}{0.53\%}  & 0.5267          & 0.5213          & 1.03\%  \\\specialrule{0.05em}{1pt}{1pt}
\multicolumn{1}{c|}{CAL}         & 0.5018          & 0.4947          & \multicolumn{1}{c|}{1.41\%}  & 0.5173          & 0.5019          & \multicolumn{1}{c|}{2.98\%}  & 0.5127          & 0.5110          & 0.33\%  \\
\multicolumn{1}{c|}{DIR}         & 0.5142          & 0.4996          & \multicolumn{1}{c|}{2.84\%}  & 0.5228          & 0.4893          & \multicolumn{1}{c|}{6.41\%}  & 0.5117          & 0.5107          & 0.20\%  \\
\multicolumn{1}{c|}{WSGNN}       & 0.5042          & 0.4872          & \multicolumn{1}{c|}{3.37\%}  & 0.5156          & 0.4978          & \multicolumn{1}{c|}{3.45\%}  & 0.5211          & 0.5178          & 0.63\%  \\\specialrule{0.05em}{1pt}{1pt}
\multicolumn{1}{c|}{MAML}        & 0.5063          & 0.5053          & \multicolumn{1}{c|}{0.20\%}  & 0.5317          & 0.5157          & \multicolumn{1}{c|}{3.01\%}  & 0.5223          & 0.5033          & 3.64\%  \\
\multicolumn{1}{c|}{ProtoNet}    & 0.5087          & 0.4893          & \multicolumn{1}{c|}{3.81\%}  & 0.5427          & 0.5200          & \multicolumn{1}{c|}{4.18\%}  & 0.5173          & 0.5003          & 3.29\%  \\\specialrule{0.05em}{1pt}{1pt}
\multicolumn{1}{c|}{Meta-GNN}    & 0.5114          & 0.4773          & \multicolumn{1}{c|}{6.67\%}  & 0.5473          & 0.5192          & \multicolumn{1}{c|}{5.13\%}  & 0.5260          & 0.4964          & 5.63\%  \\
\multicolumn{1}{c|}{GPN}         & 0.5037          & 0.4717          & \multicolumn{1}{c|}{6.35\%}  & 0.5550          & 0.5173          & \multicolumn{1}{c|}{6.79\%}  & 0.5377          & 0.5037          & 6.32\%  \\
\multicolumn{1}{c|}{G-Meta}      & 0.5556          & 0.4889          & \multicolumn{1}{c|}{12.01\%} & 0.5677          & 0.5326          & \multicolumn{1}{c|}{6.18\%}  & 0.5859*         & 0.5222          & 10.87\% \\\specialrule{0.05em}{1pt}{1pt}
\multicolumn{1}{c|}{CGFL}        & 0.5612*         & 0.5077          & \multicolumn{1}{c|}{9.53\%} & 0.6289*         & 0.6011*         & \multicolumn{1}{c|}{4.42\%}  & 0.5783          & 0.5433*         & 6.05\%  \\
\multicolumn{1}{c|}{HG-Meta}     & 0.5183          & 0.4817          & \multicolumn{1}{c|}{7.06\%}  & 0.5933          & 0.5573          & \multicolumn{1}{c|}{6.07\%}  & 0.5413          & 0.5207          & 3.81\%  \\\specialrule{0.05em}{1pt}{1pt}
\multicolumn{1}{c|}{COHF}        & \textbf{0.5889$^1$} & \textbf{0.5681} & \multicolumn{1}{c|}{3.53\%}  & \textbf{0.6392} & \textbf{0.6271} & \multicolumn{1}{c|}{1.89\%}  & \textbf{0.6014} & \textbf{0.5853} & 2.68\%  \\
\multicolumn{1}{c|}{Improvement$^2$} & \underline{4.94\%}          & \underline{8.56\%}          & \multicolumn{1}{c|}{}        & \underline{1.64\%}          & \underline{4.33\%}          & \multicolumn{1}{c|}{}        & \underline{2.65\%}          & \underline{7.73\%}          &         \\\specialrule{0.05em}{1pt}{1pt}
\multicolumn{10}{c}{\cellcolor[HTML]{E5E5E5}$\textbf{\emph{3-way 3-shot}}$} \\\specialrule{0.05em}{1pt}{1pt}
\multicolumn{1}{c|}{GAT}         & 0.3436          & 0.2669          & \multicolumn{1}{c|}{22.32\%} & 0.4778          & 0.4018          & \multicolumn{1}{c|}{15.91\%} & 0.3556          & 0.3369          & 5.26\%  \\
\multicolumn{1}{c|}{SGC}         & 0.3447          & 0.2987          & \multicolumn{1}{c|}{13.34\%} & 0.4716          & 0.3656          & \multicolumn{1}{c|}{22.48\%} & 0.3633          & 0.3309          & 8.92\%  \\
\multicolumn{1}{c|}{GIN}         & 0.3423          & 0.3049          & \multicolumn{1}{c|}{10.93\%} & 0.4916          & 0.4164          & \multicolumn{1}{c|}{15.30\%} & 0.3609          & 0.3327          & 7.81\%  \\\specialrule{0.03em}{0.7pt}{0.7pt}
\multicolumn{1}{c|}{HAN}         & 0.3612          & 0.3011          & \multicolumn{1}{c|}{16.64\%} & 0.4922          & 0.4456          & \multicolumn{1}{c|}{9.47\%} & 0.3744          & 0.3467          & 7.40\%  \\
\multicolumn{1}{c|}{MAGNN}       & 0.3581          & 0.3163          & \multicolumn{1}{c|}{11.67\%} & 0.5311          & {0.4719*} & \multicolumn{1}{c|}{11.15\%} & 0.3756          & 0.3474          & 7.51\%  \\\specialrule{0.03em}{0.7pt}{0.7pt}
\multicolumn{1}{c|}{UDA-GCN}         & 0.3412          & 0.3309         & \multicolumn{1}{c|}{3.02\%}  & 0.3356          & 0.3333          & \multicolumn{1}{c|}{0.69\%}  & 0.3392          & 0.3351          & 1.21\%  \\
\multicolumn{1}{c|}{MuSDAC}       & 0.3471          & 0.3404          & \multicolumn{1}{c|}{1.93\%}  & 0.3396          & 0.3356          & \multicolumn{1}{c|}{1.18\%}  & 0.3404          & 0.3387          & 0.50\%  \\\specialrule{0.05em}{1pt}{1pt}
\multicolumn{1}{c|}{CAL}         & 0.3413          & 0.3361          & \multicolumn{1}{c|}{1.52\%}  & 0.3598          & 0.3442          & \multicolumn{1}{c|}{4.34\%}  & 0.3469          & 0.3367          & 2.94\%  \\
\multicolumn{1}{c|}{DIR}         & 0.3467          & 0.3335          & \multicolumn{1}{c|}{3.81\%}  & 0.4044          & 0.3362          & \multicolumn{1}{c|}{16.86\%} & 0.3520          & 0.3404          & 3.30\%  \\
\multicolumn{1}{c|}{WSGNN}       & 0.3433          & 0.3404          & \multicolumn{1}{c|}{0.84\%}  & 0.3615          & 0.3319          & \multicolumn{1}{c|}{8.19\%}  & 0.3444          & 0.3356          & 2.56\%  \\\specialrule{0.03em}{0.7pt}{0.7pt}
\multicolumn{1}{c|}{MAML}        & 0.3424          & 0.3309          & \multicolumn{1}{c|}{3.36\%}  & 0.3464          & 0.3338          & \multicolumn{1}{c|}{3.64\%}  & 0.3418          & 0.3364          & 1.58\%  \\
\multicolumn{1}{c|}{ProtoNet}    & 0.3351          & 0.3333          & \multicolumn{1}{c|}{0.54\%}  & 0.4564          & 0.3484          & \multicolumn{1}{c|}{23.66\%} & 0.3384          & 0.3331          & 1.57\%  \\\specialrule{0.03em}{0.7pt}{0.7pt}
\multicolumn{1}{c|}{Meta-GNN}    & 0.3465          & 0.3274          & \multicolumn{1}{c|}{5.51\%}  & 0.4961          & 0.4123          & \multicolumn{1}{c|}{16.89\%} & 0.3692          & 0.3407          & 7.72\%  \\
\multicolumn{1}{c|}{GPN}         & 0.3476          & 0.3333          & \multicolumn{1}{c|}{4.11\%}  & 0.4413          & 0.3873          & \multicolumn{1}{c|}{12.24\%} & 0.3722          & 0.3322          & 10.75\% \\
\multicolumn{1}{c|}{G-Meta}      & 0.3667          & 0.3432          & \multicolumn{1}{c|}{6.41\%}  & 0.5189          & 0.4211          & \multicolumn{1}{c|}{18.85\%} & 0.3764          & 0.3603          & 4.28\%  \\\specialrule{0.03em}{0.7pt}{0.7pt}
\multicolumn{1}{c|}{CGFL}        & {0.3711*} & {0.3509*} & \multicolumn{1}{c|}{5.44\%}  & 0.5259          & 0.4478          & \multicolumn{1}{c|}{14.85\%} & {0.3798*} & {0.3625*} & 4.56\%  \\
\multicolumn{1}{c|}{HG-Meta}     & 0.3652          & 0.3412          & \multicolumn{1}{c|}{6.57\%}  & {0.5313*} & 0.4609          & \multicolumn{1}{c|}{13.25\%} & 0.3713          & 0.3551          & 4.36\%  \\\specialrule{0.03em}{0.7pt}{0.7pt}
\multicolumn{1}{c|}{COHF}        & \textbf{0.3842} & \textbf{0.3775} & \multicolumn{1}{c|}{1.74\%}  & \textbf{0.5417} & \textbf{0.5144} & \multicolumn{1}{c|}{5.04\%}  & \textbf{0.3915} & \textbf{0.3889} & 0.66\%  \\
\multicolumn{1}{c|}{Improvement$^2$} & \underline{3.53\%}          & \underline{7.58\%}          & \multicolumn{1}{c|}{}        & \underline{1.96\%}          & \underline{9.01\%}          & \multicolumn{1}{c|}{}        & \underline{3.08\%}          & \underline{7.28\%}          &    \\\specialrule{0.15em}{2pt}{2pt}     
\end{tabular}}}
\scriptsize{\leftline{* Result of the best-performing baseline.}}
\scriptsize{\leftline{$^1$ Results of the best-performing method are in bold.}}
\scriptsize{\leftline{$^2$ Improvements of COHF over the best-performing baseline are underlined.}}
\end{table}

\noindent \textbf{1. Inter-dataset setting:} In this setting, two different datasets from the same domain are randomly selected as the source HG and the target HG. Note that there are inherent differences between datasets within the same domain, including variations in node and relationship types, different node features, and diverse graph structures. These differences  result in multiple distribution shifts between the source HG and the target HG. Furthermore, to simulate distribution shifts between training and testing data in the target HG, we follow the approach in \cite{gui2022good} to generate training and testing data with covariate shifts based on node degrees. 

\noindent \textbf{2. Intra-dataset setting:} In this setting, both the source HG and the target HGs are derived from the same dataset. In few-shot learning, these two HGs must encompass distinct node classes (labels) for meta-training and meta-testing, respectively. This requirement mandates the selection of datasets with a substantial diversity of node classes. Consequently, we choose three out of the seven datasets containing  more than ten node classes: Douban (38 \texttt{movie} node classes), MovieLens (18 \texttt{movie} node classes), and YELP-B (16 \texttt{business} node classes). For each dataset, the classes are divided into two groups to serve as labels the source HG and the target HG, respectively. Moreover, to evaluate the performance of methods in both the I.I.D. environment and the OOD environment, we generate I.I.D. and OOD data for each dataset:
\begin{itemize}[leftmargin=*]
    \item For I.I.D. data, we randomly divide the dataset into three splits with similar sizes, which are designated as the source HG, training data, and testing data.
    \item For OOD data, we firstly follow \cite{gui2022good} to generate three splits with covariate shifts based on node degrees. Then, we introduce heterogeneity-level shifts by randomly reducing node types (excluding the target node type) in each of these three splits. These modified splits are then designated as the source HG, training data, and testing data.
\end{itemize}
We repeat the above two generation processes five times for each dataset, creating five I.I.D. and five OOD datasets. For one dataset under I.I.D. or OOD settings, we average the experimental results from its five corresponding generated datasets to derive the final results.

\noindent\textbf{Baselines.} 
Given the lack of specialized solutions for OOD generalization in HGFL problems, we select 17 representative and state-of-the-art methods as baselines, which can be adapted to our meta-learning and OOD settings with minor modifications. These baselines can be categorized into seven groups: 1) \textbf{Homogeneous GNNs:} \emph{GAT} \cite{velickovic2017graph}, \emph{SGC} \cite{wu2019simplifying} and \emph{GIN} \cite{xu2018powerful}; 2) \textbf{Heterogeneous GNNs:} \emph{HAN} \cite{wang2019heterogeneous} and \emph{MAGNN} \cite{fu2020magnn}; 3) \textbf{Unsupervised domain adaptive node classifications:} \emph{UDA-GCN} \cite{wu2020unsupervised} and \emph{MuSDAC} \cite{yang2021domain}; 4) \textbf{Graph OOD generalization methods:} \emph{CAL} \cite{sui2022causal}, \emph{DIR} \cite{fan2022debiasing}, and \emph{WSGNN} \cite{lao2022variational}; 5) \textbf{Few-shot learning methods:} \emph{MAML} \cite{finn2017model} and \emph{ProtoNet} \cite{snell2017prototypical}; 6) \textbf{Homogeneous graph few-shot learning methods:} \emph{Meta-GNN} \cite{zhou2019meta}, \emph{GPN} \cite{ding2020graph} and \emph{G-Meta} \cite{huang2020graph}; 7) \textbf{Heterogeneous graph few-shot learning methods:} \emph{CGFL} \cite{ding2023cross} and \emph{HG-Meta} \cite{zhang2022hg}. Details of these baselines are discussed in Appendix II.

\noindent\textbf{Parameter Settings.} In the $\emph{N}$-way $\emph{K}$-shot setting, $\emph{N}$ is set to \{2, 3\} and $\emph{K}$ is set to \{1, 3, 5\}. The number of tasks $\emph{m}$ is set to 100 for all datasets. To ensure fair comparisons, the embedding dimension is set to 64 for both the baselines and COHF. The baseline parameters are initially set to the values reported in the original papers and are further optimized through grid-searching to achieve optimal performance. For COHF, the subgraph $\emph{G}_\emph{s}$ is constructed by sampling 2-hop neighbors of the target node. The dimension of $\mathbf{e_2}$ ($\emph{D}$) is set to 64. In the relation encoder module, we adopt two GCNs \cite{kipf2016semi} as $\operatorname{GNN}_\emph{c}$ and $\operatorname{GNN}_\emph{u}$, ${f}^{c}_\emph{pool}(\cdot)$ is set to $\emph{max-pooling}(\cdot)$. In the multi-layer GNN module, $f^{\emph{z}_1}_\emph{pool}(\cdot)$ uses $\emph{max-pooling}(\cdot)$, and the number of GNN layers ($l$) is set to 2. In the graph learner module, $\emph{n}_\emph{att}$ is set to 8. We also adopt GCN as $\operatorname{GNN}_\emph{y}$. In the meta-learning module, the loss coefficient $\lambda$ is set to 0.4.

\noindent\textbf{Evaluation Metrics.} We adopt two widely used node classification metrics: \emph{Accuracy} and \emph{Macro-F1 score} \cite{velickovic2017graph,zhou2019meta}. To ensure a fair and accurate assessment of the performance of all methods, we perform 10 independent runs for each $\emph{N}$-way $\emph{K}$-shot setting and report the average results. 

\subsection{Experimental Results}
\noindent\textbf{Performance Comparison with Baselines (RQ1).}
Tables \ref{exp_res}-\ref{exp_acc_inter} present the performance comparison between COHF and baselines in both inter-dataset and intra-dataset settings. The results demonstrate a significant improvement of COHF over the best-performing baselines. Across all settings, COHF consistently outperforms the baselines in terms of accuracy and F1-score. Specifically, in the inter-dataset setting, COHF achieves an average improvement of 5.66\% in accuracy and 4.32\% in F1-score. In the intra-dataset setting, COHF achieves an average increase of 5.19\% in accuracy and 7.75\% in F1-score. These improvements are primarily attributed to our proposed VAE-HGNN and the node valuator module. VAE-HGNN is innovatively designed based on the SCM model and can effectively extract invariant factors in OOD environments. The node valuator module enhances knowledge transfer across HGs with distribution shifts and few-labeled data. In contrast, the baseline methods either overlook OOD environments in few-shot learning or focus solely on OOD problems within homogeneous graphs. Such limitations hinder their ability to extract generalized semantic knowledge from the source HG, leading to inferior performance when applied to a target HG with different distributions and limited labeled data.

Furthermore, it is worth noting that in Tables \ref{exp-f1} and \ref{exp_acc_inter}, almost all baseline methods demonstrate a significant drop in performance when transitioning from I.I.D. to OOD settings. This decline is especially evident in methods such as CGFL, G-Meta, and MAGNN, which perform well in I.I.D. settings. On average, there is a notable decrease of 6.28\% in accuracy and 7.72\% in F1-score across all baselines. In contrast, COHF exhibits a relatively minor average decrease of 2.8\% in accuracy and 2.16\% in F1-score. This underlines the robustness of our approach in handling distribution shifts in HGFL.

\begin{figure}[t]
\setlength{\abovecaptionskip}{0cm} 
\setlength{\belowcaptionskip}{0cm} 
\centering
\scalebox{0.37}{\includegraphics{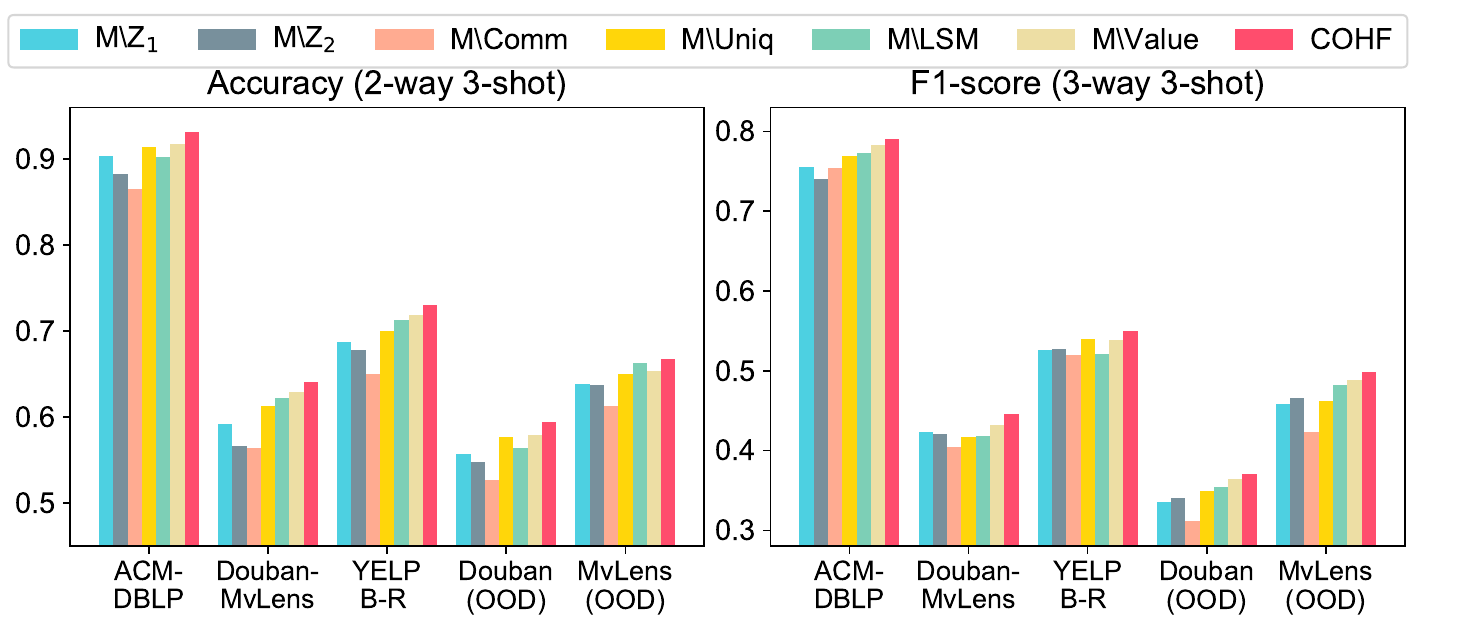}}
\caption{Node classification performance of COHF variants.}
\label{abl}
\end{figure}

\noindent\textbf{Ablation Study (RQ2).} To investigate the impact of the core modules in COHF, we create six variants focusing on three key aspects: (1) Two variants are created to explore the effectiveness of the semantic features learned based on our proposed structural causal model. The \textbf{M\textbackslash $\emph{Z}_1$} variant focuses on semantics related to $\emph{Z}_2$ by excluding $\mathbf{\bar{z}_1}$ from the input of  $\operatorname{GNN}_\emph{y}(\cdot)$. In contrast, \textbf{M\textbackslash $\emph{Z}_2$} emphasizes semantics related to $\emph{Z}_1$ by excluding $\mathbf{\bar{z}_2}$ from the input of $\operatorname{GNN}_\emph{y}(\cdot)$. (2) Two variants are created to evaluate the effectiveness of leveraging common and unique relations for OOD generalization. The \textbf{M\textbackslash Comm} variant focuses on unique relations by setting the common relation adjacency matrices $\mathbf{A}^\emph{c}$ as unit diagonal matrices. Conversely, the \textbf{M\textbackslash Uniq} variant emphasizes common relations by setting $\mathbf{A}^\emph{u}$ as unit diagonal matrices. (3) Two variants are created to study the impact of the latent space model (LSM) and node valuator module. The \textbf{M\textbackslash LSM} variant excludes the LSM and does not include structure regularization loss ($\mathcal{L}_{\emph{str}}$) when training the model. The \textbf{M\textbackslash Value} variant removes the node valuator module and averages the $\emph{K}$-shot embedded nodes within a class to derive the class prototype. From Fig. \ref{abl}, we have the following observations:

\begin{figure}[t]
\setlength{\abovecaptionskip}{0cm} 
\setlength{\belowcaptionskip}{0cm} 
\centering
\scalebox{0.356}{\includegraphics{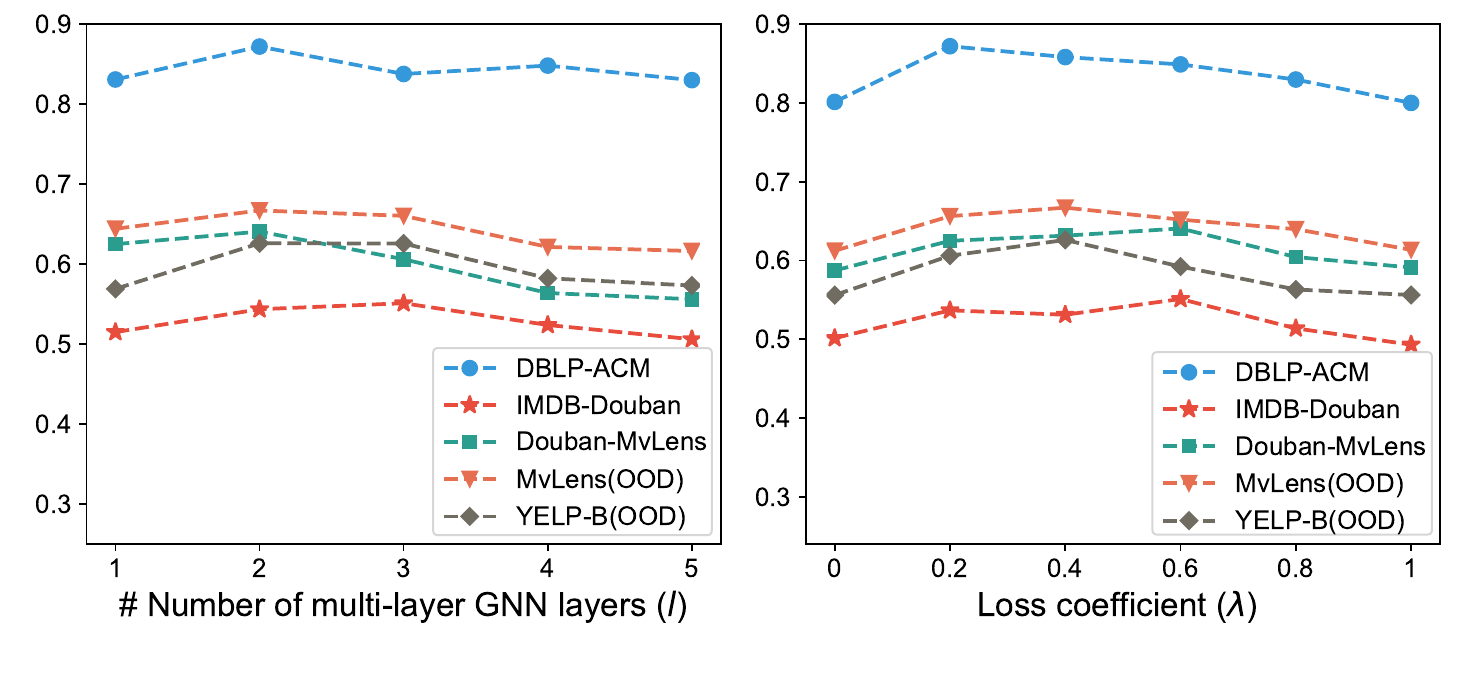}}
\caption{2-way 3-shot node classification accuracy of COHF with different parameter settings.}
\label{para}
\end{figure}

\begin{itemize}[leftmargin=*]
    \item The {M\textbackslash $\emph{Z}_1$} and {M\textbackslash $\emph{Z}_2$} variants exhibit inferior performance compared to the original COHF. This highlights the importance of the semantics associated with $\emph{Z}_1$ and $\emph{Z}_2$ in determining node labels.
    
    \item The {M\textbackslash Comm} and {M\textbackslash Uniq} variants are outperformed by COHF, indicating that adopting a relation-level perspective and utilizing common and unique relations is beneficial for identifying features that are resilient to distribution shifts.

    \item The M\textbackslash LSM and {M\textbackslash Value} variants demonstrate lower performance than COHF, emphasizing the significance of the LSM and valuator modules. The LSM module regularizes model parameters by reconstructing the graph structure, while the valuator module assesses the richness of environment-independent features of nodes in the source HG and evaluates the significance of sparsely labeled nodes in developing robust prototypes for the target HG.
\end{itemize}

\noindent\textbf{Parameter Study (RQ3).} We investigate the sensitivity of several important parameters in COHF and illustrate their impacts in Fig. \ref{para}. For the number of layers in the multi-layer GNN module, moderate values of 2 or 3 are recommended. This is mainly because meta-paths of lengths 2 and 3 are usually sufficient to capture the necessary semantic information for node classification. Meta-paths longer than these do not typically lead to further performance improvements. For the loss coefficient $\lambda$, COHF achieves optimal performance when $\lambda$ ranges between 0.2 to 0.6. Values of $\lambda$ that are either too small or too large can lead to degraded performance.

\subsection{Summary}
The experimental results demonstrate that (1) COHF considerably outperforms 17 baselines in both I.I.D. and OOD environments, achieving an average improvement of 3.03\% in accuracy and 3.25\% in F1-score for I.I.D. environments, and an average improvement of 5.43\% in accuracy and 6.04\% in F1-score for OOD environments; (2) COHF is a robust few-shot learning model capable of handling various distribution shifts in HGs. In contrast, the baseline methods cannot effectively deal with OOD environments, leading to notable performance degradation; (3) COHF exhibits superior interpretability. It is designed based on the proposed SCM. Through ablation studies that explore the effects of distribution-invariant features captured by COHF, we effectively demonstrate the efficacy of the COHF model design and the validity of the SCM. 



\section{Conclusion}
In this paper, we introduce a novel problem of OOD generalization in heterogeneous graph few-shot learning, and propose a solution called COHF. Following the invariance principle from causal theory, COHF focuses on invariant factors to address distribution shifts across HGs. Furthermore, COHF incorporates a novel value-based meta-learning framework, which facilitates learning new classes with few-labeled data and varying distributions. Extensive experiments demonstrate the superior performance of COHF. For future work, we plan to explore the application of COHF in other graph learning tasks, such as link prediction and relation mining, to exploit its potential across a wider range of contexts.

\bibliographystyle{IEEEtran}
\bibliography{reference}

\end{document}